# Improve the Autonomy of the $SE_2(3)$ group based Extended Kalman Filter for Integrated Navigation: Theoretical Analysis

JIARUI CUI

National University of Defense Technology, Changsha, Hunan 410073, China
Beijing Institute of Tracking and Communication Technology, Beijing 100094, China

MAOSONG WANG

WENQI WU

PEIQI LI

XIANFEI PAN

National University of Defense Technology, Changsha, Hunan 410073, China

Jiarui Cui was with National University of Defense Technology, Changsha, Hunan 410073, China. He is now with the Beijing Institute of Tracking and Communication Technology, Beijing 100094, China. (e-mail: cuijr1996@alumni.nudt.edu.cn).

Maosong Wang, Wenqi Wu, Peiqi Li and Xianfei Pan are with National University of Defense Technology, Changsha, Hunan 410073, China. (e-mail: wangmaosong12@nudt.edu.cn; wenqiwu_lit@hotmail.com; lpq1535623360@163.com; afeipan@126.com; ).

.

*Abstract*—One of core advantages of the $SE_2(3)$ Lie group framework for navigation modeling lies in the autonomy of error propagation. Current research on Lie group based extended Kalman filters has demonstrated that error propagation autonomy holds in low-precision applications, such as in micro electromechanical system (MEMS) based integrated navigation without considering earth rotation and inertial device biases. However, in high-precision navigation state estimation, maintaining autonomy is extremely difficult when considering with earth rotation and inertial device biases. This paper presents the theoretical analysis on the autonomy of $SE_2(3)$ group based high-precision navigation models under inertial, earth and world frame respectively. Through theoretical analysis, we find that the limitation of the traditional, trivial $SE_2(3)$ group navigation modeling method is that the presence of Coriolis force terms introduced by velocity in non-inertial frame. Therefore, a construction method for $SE_2(3)$ group navigation models is proposed, which brings the navigation models closer to full autonomy.

## I. INTRODUCTION

Work on state estimation of navigation problem dates back to the 1960s with the introduction of the extended Kalman filter (EKF) [1][7] and it was successfully applied to integrated navigation state estimation for the Apollo missions[3]. Extended Kalman filter (EKF) has essentially become the standard algorithm in industry over the past six decade [4][5][6].

Original navigation state estimation problem mainly concerns attitude estimation. The EKF is used to linearize the nonlinear attitude estimation problem on the group $SO(3)$, such as quaternions, direction cosine matrices (DCM) and Euler angles (they are used as tools to represent the 3-dimensional rotation special orthogonal group $SO(3)$ [7][8]). In EKF based navigation model, the velocity and position parameter are augmented directly to the nonlinear attitude parameter (equals to direct product group $SO(3)\times R^6$). Whereas, with the understanding of the geometry nature of rigid body dynamic in 3-demission space, the navigation states are modelled on more elegant groups, such as special Euclidean group $SE(3)$ (equals to semi-direct product group $SO(3)\ltimes R^3$) considering the nonlinear effects of attitude errors on position [9]. The $SE(3)$ has widely applied to solve pose estimation problem in the industrial and consumer-grade ground robotics[11], arial vehicles [10] and underwater vehicles[12]. The state estimation on group $SE(3)$ is naturally a nonlinear problem, thus numerous nonlinear filtering methods, such as EKF, unscented Kalman filter (UKF) [13] or particle filter (PF) [14], to simulate the nonlinear state propagation process, which have achieved better performance on position and attitude estimation to some extent. Barrua defined a novel group namely augmented special Euclidean group $SE_2(3)$ (equals to semi-direct product group $SO(3) \ltimes R^6$) that further includes velocity state and identified a group affine system based on $SE_2(3)$ [15]. This group affine system has been proven to have the linear propagation property (or autonomy property, thanks to the log linear property) of the nonlinear states in navigation problem. This led to the invariant Kalman filter (IEKF) [16] and the TFG-invariant Kalman filter (TFG-IEKF) [17] which have error propagation autonomy property in some classical navigation applications. Wang and Cui reformed the high-precision traditional integrated navigation formulations using matrix Lie group $SE_2(3)$ and proposed Lie group extended Kalman filter (LG-EKF [18]-[20]), while Chang also conducted related work [21][21]. Mahong also proposed a methodology of equivariant filter (EqF) [23][24], which provides a general design method for systems evolving in Lie group with biased input measurement and it could boil down to the invariant Kalman filter in many navigation applications practically [25]. The research on state estimation and integrated navigation on the $SE_2(3)$ group is one of the current cutting-edge topics.

The critical advantage of the Lie group $SE_2(3)$ based extended Kalman filter largely lies in the autonomy of error propagation. Autonomy is a natural property for linear systems (translational invariance), but not always true for nonlinear systems since attitude existent. By using the $SE_2(3)$ matrix group representation, the navigation state error could be represented by a compact nonlinear form [18]-[20]. This nonlinear error dynamic possesses a favorable log linear property by which arbitrary amount navigation error could be calculated exactly though nonlinear exponential map [15]. Additionally, the propagation of state error turns out to be trajectory independent that means there is no state-dependent term in the differential equation of nonlinear error. This avoids arbitrarily selection of the specific linearization point, and this is the key basis for the performance of the geometry filters (IEKF, LG-EKF, EqF) to outperform the EKF using traditional $SO(3) \times R^6$ group. However, this autonomy only holds under simplified conditions. Firstly, it must ignore sensor bias since the bias is not possible to be represented by the $SE_2(3)$ group (a common treatment is augmenting the bias by another direct product group). Secondly, it supposes a flat earth surface and no earth rotation. The curved earth surface introduces position-related gravity change term and earth rotation introduces velocity-related Coriolis force term.

At present, there are fewer systematic studies on the analysis on the error propagation autonomy. This paper revisited the $SE_2(3)$ group based navigation model under three commonly used frames (inertial frame, earth frame and world frame). According to the disruption extent on autonomy property, this paper identified two imperfect autonomy properties. Then we proposed a new construct method for the non-inertial frame navigation model by representing the velocity in the inertial frame to improve autonomy. Finally, the completely strapdown inertial navigation system (SINS)/odometer(ODO) integrated navigation system and observation equations based on the proposed method are given.

This paper mainly systematically analysis the autonomy property of $SE_2(3)$ group based EKF algorithms. In the counterpart of this paper, the performance of proposed algorithms will be evaluated by simulation and real-world navigation SINS/ODO integrated navigation application.

## II. MATHEMATICAL FOUNDATIONS

In this section, the $SE_2(3)$ group based navigation state under inertial frame, earth frame and world frames are derived in the matrix form, along with their differential equations.

### A. Lie Group based navigation state representation

The $SE_2(3)$ group is a special matrix Lie group that can be used to describe the motion of rigid bodies in the 3-dimensional space [15]. Its group elements are composed of the direction cosine matrix $C_\beta^\gamma$ on the special orthogonal group $SO(3)$, the velocity vector $v_{\alpha\beta}^\gamma$ and the position vector $r_{\alpha\beta}^\gamma$ on the 3-dimensional Euclidean space $R^3$. The $SE_2(3)$ group definition is given in (1).

$$\chi = \begin{bmatrix} C_\beta^\gamma & v_{\alpha\beta}^\gamma & r_{\alpha\beta}^\gamma - r_{\alpha\beta(0)}^\gamma \\ \mathbf{0}_{1\times 3} & 1 & 0 \\ \mathbf{0}_{1\times 3} & 0 & 1 \end{bmatrix}$$

$$\chi^{-1} = \begin{bmatrix} C_\gamma^\beta & -C_\gamma^\beta v_{\alpha\beta}^\gamma & -C_\gamma^\beta \left( r_{\alpha\beta}^\gamma - r_{\alpha\beta(0)}^\gamma \right) \\ \mathbf{0}_{1\times 3} & 1 & 0 \\ \mathbf{0}_{1\times 3} & 0 & 1 \end{bmatrix} \in SE_2(3) \quad (1)$$

The superscript and subscript of vectors $\alpha$, $\beta$ and $\gamma$ represent different coordinate system or termed frame: $\alpha$ represents the reference frame, $\beta$ the body frame, and $\gamma$ the projection frame. In this paper, the reference frame and the projection frame are selected from the inertial frame ($i$-frame), the earth centered earth fixed frame (earth frame, $e$-frame) or the world frame ($w$-frame, an earth fixed tangent plane frame). $\mathbf{0}_{m\times n}$ represents a m times n full zero matrix.

Note that all position vectors subtract the initial position to ensure they are infinitesimal at the initial time, which could generally achieve better performance for EKF algorithms [20].

## 1) *i*-frame navigation state and its derivative

The reference frame and projection frame are selected as *i*-frame, the navigation state can be compactly represented by the $SE_2(3)$ group as in (2).

$$\chi_{ib}^i = \begin{bmatrix} C_b^i & v_{ib}^i & r_{ib}^i - r_{ib(0)}^i \\ 0_{1\times 3} & 1 & 0 \\ 0_{1\times 3} & 0 & 1 \end{bmatrix} \in SE_2(3) \quad (2)$$

The reference frame is an inertial frame, whose origin is typically chosen to be the center of the earth, but can also be chosen arbitrarily. The initial position $r_{ib(0)}^i$ is deducted from the position vector $r_{ib}^i$ which points to the earth's center.

Differentiating the attitude, velocity, and position yields:

$$\begin{aligned} \dot{C}_b^i &= C_b^i (\omega_{ib}^b \times) \\ \dot{v}_{ib}^i &= C_b^i f_{ib}^b + \gamma_{ib}^i \\ \dot{r}_{ib}^i - \dot{r}_{ib(0)}^i &= v_{ib}^i \end{aligned} \quad (3)$$

where $\omega_{ib}^b$ is the angular velocity with respect to (w.r.t) the inertial space measured by the gyroscope, $f_{ib}^b$ is the specific force measured by the accelerometer, $\gamma_{ib}^i$ is the gravitational acceleration in the inertial frame, and $\times$ represents a skew-symmetric matrix formed by a vector.

The differential equations of navigation state (attitude, velocity, and position) in (3) can be expressed in matrix form in (4).

$$\frac{d}{dt}(\chi_{ib}^i) = \chi_{ib}^i \begin{bmatrix} (\omega_{ib}^b \times) & f_{ib}^b & 0_{3\times 1} \\ 0_{1\times 3} & 0 & 1 \\ 0_{1\times 3} & 0 & 0 \end{bmatrix} + \begin{bmatrix} 0_{3\times 3} & \gamma_{ib}^i & 0_{3\times 1} \\ 0_{1\times 3} & 0 & -1 \\ 0_{1\times 3} & 0 & 0 \end{bmatrix} \chi_{ib}^i \quad (4)$$

$$\triangleq \chi_{ib}^i W_1^i + W_2^i \chi_{ib}^i$$

## 2) *e*-frame navigation state and its derivative

The reference frame and projection frame are selected as *e*-frame, the navigation states can be represented as in (5).

$$\chi_{eb}^e = \begin{bmatrix} C_b^e & v_{eb}^e & r_{eb}^e - r_{eb(0)}^e \\ 0_{1\times 3} & 1 & 0 \\ 0_{1\times 3} & 0 & 1 \end{bmatrix} \in SE_2(3) \quad (5)$$

where $r_{eb(0)}^e$ denotes the initial position in the *e*-frame.

Differential equations of navigation states are given in (6).

$$\begin{aligned} \dot{C}_b^e &= C_b^e (\omega_{ib}^b \times) - (\omega_{ie}^e \times) C_b^e \\ \dot{v}_{eb}^e &= C_b^e f_{ib}^b - 2(\omega_{ie}^e \times) v_{eb}^e + g_{eb}^e \\ \dot{r}_{eb}^e - \dot{r}_{eb(0)}^e &= \dot{r}_{eb}^e = v_{eb}^e \end{aligned} \quad (6)$$

Rewrite (6) in a matrix form as shown in (7):

$$\frac{d}{dt}(\chi_{eb}^e) = \chi_{eb}^e \begin{bmatrix} (\omega_{ib}^b \times) & f_{ib}^b & 0_{3\times 1} \\ 0_{1\times 3} & 0 & 1 \\ 0_{1\times 3} & 0 & 0 \end{bmatrix} + \begin{bmatrix} -(\omega_{ie}^e \times) & g_{eb}^e & 0_{3\times 1} \\ 0_{1\times 3} & 0 & -1 \\ 0_{1\times 3} & 0 & 0 \end{bmatrix} \chi_{eb}^e$$

$$+ \begin{bmatrix} -(\omega_{ie}^e \times) & 0_{3\times 1} & 0_{3\times 1} \\ 0_{1\times 3} & 0 & 0 \\ 0_{1\times 3} & 0 & 0 \end{bmatrix} \chi_{eb}^e \begin{bmatrix} 0_{3\times 3} & 0_{3\times 1} & 0_{3\times 1} \\ 0_{1\times 3} & 1 & 0 \\ 0_{1\times 3} & 0 & -1 \end{bmatrix} \quad (7)$$

$$\triangleq \chi_{eb}^e W_1^e + W_2^e \chi_{eb}^e + W_3^e \chi_{eb}^e W_4^e$$

where $\omega_{ie}^e$ is the earth rotation angular velocity with respect to the inertial space represented in the *e*-frame, $g_{eb}^e$ is the gravity acceleration in the *e*-frame.

## 3) *w*-frame navigation state and its derivative

The reference frame and projection frame are selected as *w*-frame, the navigation state can be represented as in (8).

$$\chi_{wb}^w = \begin{bmatrix} C_b^w & v_{wb}^w & r_{wb}^w - r_{wb(0)}^w \\ 0_{1\times 3} & 1 & 0 \\ 0_{1\times 3} & 0 & 1 \end{bmatrix} \in SE_2(3) \quad (8)$$

where $r_{wb(0)}^w$ denotes the initial position in the *w*-frame. Differential equations of navigation state are given in (9).

$$\begin{aligned} \dot{C}_b^w &= C_b^w (\omega_{ib}^b \times) - (\omega_{ie}^w \times) C_b^w \\ \dot{v}_{wb}^w &= C_b^w f_{ib}^b - 2(\omega_{ie}^w \times) v_{wb}^w + g_{wb}^w \\ \dot{r}_{wb}^w - \dot{r}_{wb(0)}^w &= \dot{r}_{wb}^w = v_{wb}^w \end{aligned} \quad (9)$$

where $\omega_{ie}^w$ is the earth rotation angular velocity with respect to the inertial space represented in the *w*-frame, $g_{eb}^w$ is the gravity acceleration in the *e*-frame.

Rewrite (9) in a matrix form as follows as shown in (10).

$$\frac{d}{dt}(\chi_{wb}^w) = \chi_{wb}^w \begin{bmatrix} (\omega_{ib}^b \times) & f_{ib}^b & 0_{3\times 1} \\ 0_{1\times 3} & 0 & 1 \\ 0_{1\times 3} & 0 & 0 \end{bmatrix} + \begin{bmatrix} -(\omega_{ie}^w \times) & g_{eb}^w & 0_{3\times 1} \\ 0_{1\times 3} & 0 & -1 \\ 0_{1\times 3} & 0 & 0 \end{bmatrix} \chi_{wb}^w$$

$$+ \begin{bmatrix} -(\omega_{ie}^w \times) & 0_{3\times 1} & 0_{3\times 1} \\ 0_{1\times 3} & 0 & 0 \\ 0_{1\times 3} & 0 & 0 \end{bmatrix} \chi_{wb}^w \begin{bmatrix} 0_{3\times 3} & 0_{3\times 1} & 0_{3\times 1} \\ 0_{1\times 3} & 1 & 0 \\ 0_{1\times 3} & 0 & -1 \end{bmatrix} \quad (10)$$

$$\triangleq \chi_{wb}^w W_1^w + W_2^w \chi_{wb}^w + W_3^w \chi_{wb}^w W_4^w$$

It can be seen that equation (7) and (10) are very similar, differing only in projection frame. However, there exists a performance difference between them [20].

### B. Lie Group based navigation error state

The navigation error state is also an element on $SE_2(3)$ group, which is defined as the difference between the true state $\chi$ and the estimated state $\tilde{\chi}$. Due to the non-commutativity of matrix products, there are right multiplication and left multiplication error definition. The corresponding left error $\eta^L$ and right error $\eta^R$ are defined as follow.
$$\boldsymbol{\eta}^R = \boldsymbol{\chi}\tilde{\boldsymbol{\chi}}^{-1}, \quad \boldsymbol{\eta}^L = \tilde{\boldsymbol{\chi}}^{-1}\boldsymbol{\chi} \quad (11)$$

The navigation error state can also be defined as follows.
$$\boldsymbol{\eta}^R = \tilde{\boldsymbol{\chi}}\boldsymbol{\chi}^{-1}, \quad \boldsymbol{\eta}^L = \boldsymbol{\chi}^{-1}\tilde{\boldsymbol{\chi}} \quad (12)$$

The errors defined in (11) and (12) are inverse matrices of each other. They can both be used if the correct group multiplication is conducted. In this paper, definition (11) is used as in our previous works [18]-[20].

Derivate (11) we get (13) and (14)
$$\frac{d}{dt}\boldsymbol{\eta}^R = \left(\frac{d}{dt}\boldsymbol{\chi}\right)\tilde{\boldsymbol{\chi}}^{-1} - \boldsymbol{\eta}^R\left(\frac{d}{dt}\tilde{\boldsymbol{\chi}}\right)\tilde{\boldsymbol{\chi}}^{-1} \quad (13)$$

$$\frac{d}{dt}\boldsymbol{\eta}^L = \tilde{\boldsymbol{\chi}}^{-1}\left(\frac{d}{dt}\boldsymbol{\chi}\right) - \tilde{\boldsymbol{\chi}}^{-1}\left(\frac{d}{dt}\tilde{\boldsymbol{\chi}}\right)\boldsymbol{\eta}^L \quad (14)$$

Rewrite (11) and give its derivative as follows
$$\boldsymbol{\chi} = \boldsymbol{\eta}^R\tilde{\boldsymbol{\chi}}, \quad \boldsymbol{\chi} = \tilde{\boldsymbol{\chi}}\boldsymbol{\eta}^L \quad (15)$$

$$\frac{d}{dt}\boldsymbol{\chi} = \frac{d}{dt}(\boldsymbol{\eta}^R\tilde{\boldsymbol{\chi}}) = \boldsymbol{\eta}^R\left(\frac{d}{dt}\tilde{\boldsymbol{\chi}}\right) + \left(\frac{d}{dt}\boldsymbol{\eta}^R\right)\tilde{\boldsymbol{\chi}} \quad (16)$$

$$\frac{d}{dt}\boldsymbol{\chi} = \frac{d}{dt}(\tilde{\boldsymbol{\chi}}\boldsymbol{\eta}^L) = \left(\frac{d}{dt}\tilde{\boldsymbol{\chi}}\right)\boldsymbol{\eta}^L + \tilde{\boldsymbol{\chi}}\left(\frac{d}{dt}\boldsymbol{\eta}^L\right) \quad (17)$$

Equation (13)-(17) are helpful to subsequential derivations.

1) *i-frame navigation error state and its derivative*
Taking the *i*-frame as the reference frame and projection frame, the right error can be written as (18).

$$\boldsymbol{\eta}_{ib}^{iR} = \boldsymbol{\chi}_{ib}^i(\tilde{\boldsymbol{\chi}}_{ib}^i)^{-1} = \begin{bmatrix} \boldsymbol{C}_b^i & \boldsymbol{v}_{ib}^i & \boldsymbol{r}_{ib}^i - \boldsymbol{r}_{ib(0)}^i \\ \boldsymbol{0}_{1\times 3} & 1 & 0 \\ \boldsymbol{0}_{1\times 3} & 0 & 1 \end{bmatrix} \begin{bmatrix} \tilde{\boldsymbol{C}}_b^i & \tilde{\boldsymbol{v}}_{ib}^i & \tilde{\boldsymbol{r}}_{ib}^i - \boldsymbol{r}_{ib(0)}^i \\ \boldsymbol{0}_{1\times 3} & 1 & 0 \\ \boldsymbol{0}_{1\times 3} & 0 & 1 \end{bmatrix}^{-1}$$

$$= \begin{bmatrix} \boldsymbol{C}_{i'}^i & \boldsymbol{v}_{ib}^i - \tilde{\boldsymbol{v}}_{ib}^i + (\boldsymbol{I}_3 - \boldsymbol{C}_{i'}^i)\tilde{\boldsymbol{v}}_{ib}^i & \boldsymbol{r}_{ib}^i - \tilde{\boldsymbol{r}}_{ib}^i + (\boldsymbol{I}_3 - \boldsymbol{C}_{i'}^i)(\tilde{\boldsymbol{r}}_{ib}^i - \boldsymbol{r}_{ib(0)}^i) \\ \boldsymbol{0}_{1\times 3} & 1 & 0 \\ \boldsymbol{0}_{1\times 3} & 0 & 1 \end{bmatrix} \quad (18)$$

$$\approx \begin{bmatrix} \exp(\boldsymbol{\phi}_{ib}^i\times) & -\delta\boldsymbol{v}_{ib}^i + (\tilde{\boldsymbol{v}}_{ib}^i\times)\boldsymbol{\phi}_{ib}^i & -\delta\boldsymbol{r}_{ib}^i + (\tilde{\boldsymbol{r}}_{ib}^i - \boldsymbol{r}_{ib(0)}^i)\times\boldsymbol{\phi}_{ib}^i \\ \boldsymbol{0}_{1\times 3} & 1 & 0 \\ \boldsymbol{0}_{1\times 3} & 0 & 1 \end{bmatrix}$$

where $\tilde{\boldsymbol{r}}_{ib}^i = \boldsymbol{r}_{ib}^i + \delta\boldsymbol{r}_{ib}^i$ and $\tilde{\boldsymbol{v}}_{ib}^i = \boldsymbol{v}_{ib}^i + \delta\boldsymbol{v}_{ib}^i$ are the position and velocity containing error $\delta\boldsymbol{r}_{ib}^i$ and $\delta\boldsymbol{v}_{ib}^i$. The attitude error is expressed as $\boldsymbol{C}_{i'}^i$ which is the DCM from the estimated *i*-frame (*i'*-frame) to the true *i*-frame. It satisfies $\boldsymbol{C}_{i'}^i = \boldsymbol{C}_b^i\tilde{\boldsymbol{C}}_i^b = \exp(\boldsymbol{\phi}_{ib}^i\times) \approx \boldsymbol{I}_3 + (\boldsymbol{\phi}_{ib}^i\times)$. The exp represents matrix exponential function, and $\boldsymbol{\phi}_{ib}^i$ is the attitude error expressed in the *i*-frame.

The left error in *i*-frame is shown in (19).

$$\boldsymbol{\eta}_{ib}^{bL} = (\tilde{\boldsymbol{\chi}}_{ib}^i)^{-1}\boldsymbol{\chi}_{ib}^i = \begin{bmatrix} \tilde{\boldsymbol{C}}_b^i & \tilde{\boldsymbol{v}}_{ib}^i & \tilde{\boldsymbol{r}}_{ib}^i - \boldsymbol{r}_{ib(0)}^i \\ \boldsymbol{0}_{1\times 3} & 1 & 0 \\ \boldsymbol{0}_{1\times 3} & 0 & 1 \end{bmatrix}^{-1} \begin{bmatrix} \boldsymbol{C}_b^i & \boldsymbol{v}_{ib}^i & \boldsymbol{r}_{ib}^i - \boldsymbol{r}_{ib(0)}^i \\ \boldsymbol{0}_{1\times 3} & 1 & 0 \\ \boldsymbol{0}_{1\times 3} & 0 & 1 \end{bmatrix}$$

$$= \begin{bmatrix} \boldsymbol{C}_i^{b'}\boldsymbol{C}_b^i & \boldsymbol{C}_i^{b'}\boldsymbol{v}_{ib}^i - \boldsymbol{C}_i^{b'}\tilde{\boldsymbol{v}}_{ib}^i & \boldsymbol{C}_i^{b'}(\boldsymbol{r}_{ib}^i - \boldsymbol{r}_{ib(0)}^i) - \boldsymbol{C}_i^{b'}(\tilde{\boldsymbol{r}}_{ib}^i - \boldsymbol{r}_{ib(0)}^i) \\ \boldsymbol{0}_{1\times 3} & 1 & 0 \\ \boldsymbol{0}_{1\times 3} & 0 & 1 \end{bmatrix} \quad (19)$$

$$\approx \begin{bmatrix} \exp(-\boldsymbol{\phi}_{ib}^b\times) & -\delta\boldsymbol{v}_{ib}^b & -\delta\boldsymbol{r}_{ib}^b \\ \boldsymbol{0}_{1\times 3} & 1 & 0 \\ \boldsymbol{0}_{1\times 3} & 0 & 1 \end{bmatrix}$$

where $\boldsymbol{C}_b^{b'}$ is the attitude error of the left error which is the DCM from the *b*-frame to the estimated *b*-frame (*b'*-frame). It satisfies $\boldsymbol{C}_b^{b'} = \tilde{\boldsymbol{C}}_i^b\boldsymbol{C}_b^i = \exp(-\boldsymbol{\phi}_{ib}^b\times) \approx \boldsymbol{I}_3 - (\boldsymbol{\phi}_{ib}^b\times)$ and $\boldsymbol{\phi}_{ib}^b$ is the attitude error vector expressed in the body frame.

According to equation (4) and (13), the differential equation of right error in inertial frame can be derived as:
$$\frac{d}{dt}\boldsymbol{\eta}_{ib}^{iR} = \left(\frac{d}{dt}\boldsymbol{\chi}_{ib}^i\right)(\tilde{\boldsymbol{\chi}}_{ib}^i)^{-1} - \boldsymbol{\eta}_{ib}^{iR}\left(\frac{d}{dt}\tilde{\boldsymbol{\chi}}_{ib}^i\right)(\tilde{\boldsymbol{\chi}}_{ib}^i)^{-1} \quad (20)$$

Expanding and simplifying the above equation using (1), (2) and (4), we get the differential equation of right error in matrix form as shown in (21).

$$\frac{d}{dt}\boldsymbol{\eta}_{ib}^{iR} = \begin{bmatrix} \boldsymbol{0}_{3\times 3} & \boldsymbol{\gamma}_{ib}^i & \boldsymbol{0}_{3\times 1} \\ \boldsymbol{0}_{1\times 3} & 0 & -1 \\ \boldsymbol{0}_{1\times 3} & 0 & 0 \end{bmatrix}\boldsymbol{\eta}_{ib}^{iR} - \boldsymbol{\eta}_{ib}^{iR}\begin{bmatrix} \boldsymbol{0}_{3\times 3} & \tilde{\boldsymbol{\gamma}}_{ib}^i & \boldsymbol{0}_{3\times 1} \\ \boldsymbol{0}_{1\times 3} & 0 & -1 \\ \boldsymbol{0}_{1\times 3} & 0 & 0 \end{bmatrix}$$

$$-\boldsymbol{\eta}_{ib}^{iR}\tilde{\boldsymbol{\chi}}_{ib}^i\begin{bmatrix} (\delta\boldsymbol{\omega}_{ib}^b\times) & \delta\boldsymbol{f}_{ib}^b & \boldsymbol{0}_{3\times 1} \\ \boldsymbol{0}_{1\times 3} & 0 & 0 \\ \boldsymbol{0}_{1\times 3} & 0 & 0 \end{bmatrix}(\tilde{\boldsymbol{\chi}}_{ib}^i)^{-1} \quad (21)$$

Similarly, the differential equation of left error is given in (22).
$$\frac{d}{dt}\boldsymbol{\eta}_{ib}^{bL} = (\tilde{\boldsymbol{\chi}}_{ib}^i)^{-1}\left(\frac{d}{dt}\boldsymbol{\chi}_{ib}^i\right) - (\tilde{\boldsymbol{\chi}}_{ib}^i)^{-1}\left(\frac{d}{dt}\tilde{\boldsymbol{\chi}}_{ib}^i\right)\boldsymbol{\eta}_{ib}^{bL} \quad (22)$$

Expanding and simplifying the above equation using (1), (2) and (4), we get the differential equation of left error in matrix form as shown in (23).

$$\frac{d}{dt}\boldsymbol{\eta}_{ib}^{bL} = \boldsymbol{\eta}_{ib}^{bL}\begin{bmatrix} (\boldsymbol{\omega}_{ib}^b\times) & \boldsymbol{f}_{ib}^b & \boldsymbol{0}_{3\times 1} \\ \boldsymbol{0}_{1\times 3} & 0 & 1 \\ \boldsymbol{0}_{1\times 3} & 0 & 0 \end{bmatrix} - \begin{bmatrix} (\tilde{\boldsymbol{\omega}}_{ib}^b\times) & \tilde{\boldsymbol{f}}_{ib}^b & \boldsymbol{0}_{3\times 1} \\ \boldsymbol{0}_{1\times 3} & 0 & 1 \\ \boldsymbol{0}_{1\times 3} & 0 & 0 \end{bmatrix}\boldsymbol{\eta}_{ib}^{bL}$$

$$-(\tilde{\boldsymbol{\chi}}_{ib}^i)^{-1}\begin{bmatrix} \boldsymbol{0}_{3\times 3} & \delta\boldsymbol{\gamma}_{ib}^i & \boldsymbol{0}_{3\times 1} \\ \boldsymbol{0}_{1\times 3} & 0 & 0 \\ \boldsymbol{0}_{1\times 3} & 0 & 0 \end{bmatrix}\tilde{\boldsymbol{\chi}}_{ib}^i\boldsymbol{\eta}_{ib}^{bL} \quad (23)$$

2) *e-frame navigation error state and its derivative*
Taking the *e*-frame as the reference frame and projection frame, the right error can be written as (24).

$$\boldsymbol{\eta}_{eb}^{eR} = \boldsymbol{\chi}_{eb}^{e}\left(\tilde{\boldsymbol{\chi}}_{eb}^{e}\right)^{-1} = \begin{bmatrix} \boldsymbol{C}_{b}^{e} & \boldsymbol{v}_{eb}^{e} & \boldsymbol{r}_{eb}^{e} - \boldsymbol{r}_{eb(0)}^{e} \\ \boldsymbol{0}_{1\times 3} & 1 & 0 \\ \boldsymbol{0}_{1\times 3} & 0 & 1 \end{bmatrix} \begin{bmatrix} \tilde{\boldsymbol{C}}_{b}^{e} & \tilde{\boldsymbol{v}}_{eb}^{e} & \tilde{\boldsymbol{r}}_{eb}^{e} - \boldsymbol{r}_{eb(0)}^{e} \\ \boldsymbol{0}_{1\times 3} & 1 & 0 \\ \boldsymbol{0}_{1\times 3} & 0 & 1 \end{bmatrix}^{-1}$$

$$\approx \begin{bmatrix} \exp(\boldsymbol{\phi}_{eb}^{e}) & -\delta\boldsymbol{v}_{eb}^{e} + (\tilde{\boldsymbol{v}}_{eb}^{e}\times)\boldsymbol{\phi}_{eb}^{e} & -\delta\boldsymbol{r}_{eb}^{e} + (\tilde{\boldsymbol{r}}_{eb}^{e} - \boldsymbol{r}_{eb(0)}^{e})\times\boldsymbol{\phi}_{eb}^{e} \\ \boldsymbol{0}_{1\times 3} & 1 & 0 \\ \boldsymbol{0}_{1\times 3} & 0 & 1 \end{bmatrix} \quad (24)$$

where $\boldsymbol{C}_{e'}^{e}$ is the attitude error of right error which is the DCM from the estimated $e$-frame ($e'$-frame) to the true $e$-frame. It satisfies $\boldsymbol{C}_{e'}^{e} = \boldsymbol{C}_{b}^{e}\tilde{\boldsymbol{C}}_{e}^{b} = \exp(\boldsymbol{\phi}_{eb}^{e}\times) \approx \boldsymbol{I}_{3} + (\boldsymbol{\phi}_{eb}^{e}\times)$. $\boldsymbol{\phi}_{eb}^{e}$ is the attitude error expressed in the $e$-frame.

Taking the $e$-frame as the reference frame and the $b$-frame as the projection frame, the left error is expressed as in (25).

$$\boldsymbol{\eta}_{eb}^{bL} = \left(\tilde{\boldsymbol{\chi}}_{eb}^{e}\right)^{-1}\boldsymbol{\chi}_{eb}^{e} = \begin{bmatrix} \tilde{\boldsymbol{C}}_{b}^{e} & \tilde{\boldsymbol{v}}_{eb}^{e} & \tilde{\boldsymbol{r}}_{eb}^{e} - \boldsymbol{r}_{eb(0)}^{e} \\ \boldsymbol{0}_{1\times 3} & 1 & 0 \\ \boldsymbol{0}_{1\times 3} & 0 & 1 \end{bmatrix}^{-1} \begin{bmatrix} \boldsymbol{C}_{b}^{e} & \boldsymbol{v}_{eb}^{e} & \boldsymbol{r}_{eb}^{e} - \boldsymbol{r}_{eb(0)}^{e} \\ \boldsymbol{0}_{1\times 3} & 1 & 0 \\ \boldsymbol{0}_{1\times 3} & 0 & 1 \end{bmatrix}$$

$$\approx \begin{bmatrix} \exp(-\boldsymbol{\phi}_{eb}^{b}\times) & -\delta\boldsymbol{v}_{eb}^{b} & -\delta\boldsymbol{r}_{eb}^{b} \\ \boldsymbol{0}_{1\times 3} & 1 & 0 \\ \boldsymbol{0}_{1\times 3} & 0 & 1 \end{bmatrix} \quad (25)$$

The attitude error of the left error is expressed as $\boldsymbol{C}_{b}^{b'}$ which is the DCM from the true $b$-frame to the estimated $b$-frame ($b'$-frame). It satisfies $\boldsymbol{C}_{b}^{b'} \approx \boldsymbol{I}_{3} - (\boldsymbol{\phi}_{eb}^{b}\times)$ and $\boldsymbol{\phi}_{eb}^{b}$ is the attitude error expressed in the body frame.

According to equation (4) and (13), the differential equation of right error state in $e$-frame can be derived as:

$$\frac{d}{dt}\boldsymbol{\eta}_{eb}^{eR} = \left(\frac{d}{dt}\boldsymbol{\chi}_{eb}^{e}\right)\left(\tilde{\boldsymbol{\chi}}_{eb}^{e}\right)^{-1} - \boldsymbol{\eta}_{eb}^{eR}\left(\frac{d}{dt}\tilde{\boldsymbol{\chi}}_{eb}^{e}\right)\left(\tilde{\boldsymbol{\chi}}_{eb}^{e}\right)^{-1} \quad (26)$$

Expanding and simplifying the above equation using (1), (5) and (7), we get the differential equation of right error in matrix form as shown in (27).

$$\frac{d}{dt}\boldsymbol{\eta}_{eb}^{eR} = \left\{\begin{bmatrix} -(\boldsymbol{\omega}_{ie}^{e}\times) & \boldsymbol{g}_{eb}^{e} & \boldsymbol{0}_{3\times 1} \\ \boldsymbol{0}_{1\times 3} & 0 & -1 \\ \boldsymbol{0}_{1\times 3} & 0 & 0 \end{bmatrix}\boldsymbol{\eta}_{eb}^{eR} - \boldsymbol{\eta}_{eb}^{eR}\tilde{\boldsymbol{\chi}}_{eb}^{e}\begin{bmatrix} (\delta\boldsymbol{\omega}_{ib}^{b}\times) & \delta\boldsymbol{f}_{ib}^{b} & \boldsymbol{0}_{3\times 1} \\ \boldsymbol{0}_{1\times 3} & 0 & 0 \\ \boldsymbol{0}_{1\times 3} & 0 & 0 \end{bmatrix}\left(\tilde{\boldsymbol{\chi}}_{eb}^{e}\right)^{-1}\right\}$$

$$+ \left\{\begin{bmatrix} -(\boldsymbol{\omega}_{ie}^{e}\times) & \boldsymbol{0}_{3\times 1} & \boldsymbol{0}_{3\times 1} \\ \boldsymbol{0}_{1\times 3} & 0 & 0 \\ \boldsymbol{0}_{1\times 3} & 0 & 0 \end{bmatrix}\boldsymbol{\eta}_{eb}^{eR} - \boldsymbol{\eta}_{eb}^{eR}\tilde{\boldsymbol{\chi}}_{eb}^{e}\begin{bmatrix} \boldsymbol{0}_{3\times 3} & \boldsymbol{0}_{3\times 1} & \boldsymbol{0}_{3\times 1} \\ \boldsymbol{0}_{1\times 3} & 1 & 0 \\ \boldsymbol{0}_{1\times 3} & 0 & -1 \end{bmatrix}\left(\tilde{\boldsymbol{\chi}}_{eb}^{e}\right)^{-1}\right\} \quad (27)$$

Similarly, we get the differential equation of left error as shown in (28).

$$\frac{d}{dt}\boldsymbol{\eta}_{eb}^{bL} = \left(\tilde{\boldsymbol{\chi}}_{eb}^{e}\right)^{-1}\left(\frac{d}{dt}\boldsymbol{\chi}_{eb}^{e}\right) - \left(\tilde{\boldsymbol{\chi}}_{eb}^{e}\right)^{-1}\left(\frac{d}{dt}\tilde{\boldsymbol{\chi}}_{eb}^{e}\right)\boldsymbol{\eta}_{eb}^{bL}$$

$$= \boldsymbol{\eta}_{eb}^{bL}\begin{bmatrix} (\boldsymbol{\omega}_{ib}^{b}\times) & \boldsymbol{f}_{ib}^{b} & \boldsymbol{0}_{3\times 1} \\ \boldsymbol{0}_{1\times 3} & 0 & 1 \\ \boldsymbol{0}_{1\times 3} & 0 & 0 \end{bmatrix} - \begin{bmatrix} (\tilde{\boldsymbol{\omega}}_{ib}^{b}\times) & \tilde{\boldsymbol{f}}_{ib}^{b} & \boldsymbol{0}_{3\times 1} \\ \boldsymbol{0}_{1\times 3} & 0 & 1 \\ \boldsymbol{0}_{1\times 3} & 0 & 0 \end{bmatrix}\boldsymbol{\eta}_{eb}^{bL}$$

$$-\left(\tilde{\boldsymbol{\chi}}_{eb}^{e}\right)^{-1}\begin{bmatrix} \boldsymbol{0}_{3\times 3} & \delta\boldsymbol{g}_{eb}^{e} & \boldsymbol{0}_{3\times 1} \\ \boldsymbol{0}_{1\times 3} & 0 & -1 \\ \boldsymbol{0}_{1\times 3} & 0 & 0 \end{bmatrix}\boldsymbol{\chi}_{eb}^{e} \quad (28)$$

$$+\left(\tilde{\boldsymbol{\chi}}_{eb}^{e}\right)^{-1}\left\{\begin{bmatrix} -(\boldsymbol{\omega}_{ie}^{e}\times) & \boldsymbol{0}_{3\times 1} & \boldsymbol{0}_{3\times 1} \\ \boldsymbol{0}_{1\times 3} & 0 & 0 \\ \boldsymbol{0}_{1\times 3} & 0 & 0 \end{bmatrix}\tilde{\boldsymbol{\chi}}_{eb}^{e}\boldsymbol{\eta}_{eb}^{bL}\begin{bmatrix} \boldsymbol{0}_{3\times 3} & \boldsymbol{0}_{3\times 1} & \boldsymbol{0}_{3\times 1} \\ \boldsymbol{0}_{1\times 3} & 1 & 0 \\ \boldsymbol{0}_{1\times 3} & 0 & -1 \end{bmatrix} - \begin{bmatrix} \boldsymbol{0}_{3\times 3} & \boldsymbol{0}_{3\times 1} & \boldsymbol{0}_{3\times 1} \\ \boldsymbol{0}_{1\times 3} & 1 & 0 \\ \boldsymbol{0}_{1\times 3} & 0 & -1 \end{bmatrix}\boldsymbol{\eta}_{eb}^{bL}\right\}$$

### 3) $w$-frame navigation error state and its derivative

Taking the $w$-frame as the reference frame and projection frame, can be written as (29).

$$\boldsymbol{\eta}_{wb}^{wR} = \boldsymbol{\chi}_{wb}^{w}\left(\tilde{\boldsymbol{\chi}}_{wb}^{w}\right)^{-1} = \begin{bmatrix} \boldsymbol{C}_{b}^{w} & \boldsymbol{v}_{wb}^{w} & \boldsymbol{r}_{wb}^{w} - \boldsymbol{r}_{wb(0)}^{w} \\ \boldsymbol{0}_{1\times 3} & 1 & 0 \\ \boldsymbol{0}_{1\times 3} & 0 & 1 \end{bmatrix}\begin{bmatrix} \tilde{\boldsymbol{C}}_{b}^{w} & \tilde{\boldsymbol{v}}_{wb}^{w} & \tilde{\boldsymbol{r}}_{wb}^{w} - \boldsymbol{r}_{wb(0)}^{w} \\ \boldsymbol{0}_{1\times 3} & 1 & 0 \\ \boldsymbol{0}_{1\times 3} & 0 & 1 \end{bmatrix}^{-1}$$

$$\approx \begin{bmatrix} \exp(\boldsymbol{\phi}_{wb}^{w}) & -\delta\boldsymbol{v}_{wb}^{w} + (\tilde{\boldsymbol{v}}_{wb}^{w}\times)\boldsymbol{\phi}_{wb}^{w} & -\delta\boldsymbol{r}_{wb}^{w} + (\tilde{\boldsymbol{r}}_{wb}^{w} - \boldsymbol{r}_{wb(0)}^{w})\times\boldsymbol{\phi}_{wb}^{w} \\ \boldsymbol{0}_{1\times 3} & 1 & 0 \\ \boldsymbol{0}_{1\times 3} & 0 & 1 \end{bmatrix} \quad (29)$$

where $\boldsymbol{C}_{w'}^{w}$ is the DCM from the estimated $w$-frame ($w'$-frame) to the true $w$-frame. It satisfies $\boldsymbol{C}_{w'}^{w} = \boldsymbol{C}_{b}^{w}\tilde{\boldsymbol{C}}_{w}^{b} = \exp(\boldsymbol{\phi}_{wb}^{w}\times) \approx \boldsymbol{I}_{3} + (\boldsymbol{\phi}_{wb}^{w}\times)$. $\boldsymbol{\phi}_{wb}^{w}$ is the attitude error expressed in the world frame.

Taking the $w$-frame the reference frame and the $b$-frame the projection frame, the left error is given in (30).

$$\boldsymbol{\eta}_{wb}^{bL} = \left(\tilde{\boldsymbol{\chi}}_{wb}^{w}\right)^{-1}\boldsymbol{\chi}_{wb}^{w} = \begin{bmatrix} \tilde{\boldsymbol{C}}_{b}^{w} & \tilde{\boldsymbol{v}}_{wb}^{w} & \tilde{\boldsymbol{r}}_{wb}^{w} - \boldsymbol{r}_{wb(0)}^{w} \\ \boldsymbol{0}_{1\times 3} & 1 & 0 \\ \boldsymbol{0}_{1\times 3} & 0 & 1 \end{bmatrix}^{-1}\begin{bmatrix} \boldsymbol{C}_{b}^{w} & \boldsymbol{v}_{wb}^{w} & \boldsymbol{r}_{wb}^{w} - \boldsymbol{r}_{wb(0)}^{w} \\ \boldsymbol{0}_{1\times 3} & 1 & 0 \\ \boldsymbol{0}_{1\times 3} & 0 & 1 \end{bmatrix}$$

$$\approx \begin{bmatrix} \exp(-\boldsymbol{\phi}_{wb}^{b}\times) & -\delta\boldsymbol{v}_{wb}^{b} & -\delta\boldsymbol{r}_{wb}^{b} \\ \boldsymbol{0}_{1\times 3} & 1 & 0 \\ \boldsymbol{0}_{1\times 3} & 0 & 1 \end{bmatrix}$$

(30)

where $\boldsymbol{C}_{b}^{b'}$ is the attitude error of the left error and it verifies $\boldsymbol{C}_{b}^{b'} = \tilde{\boldsymbol{C}}_{w}^{b}\boldsymbol{C}_{b}^{w} = \exp(-\boldsymbol{\phi}_{wb}^{b}\times) \approx \boldsymbol{I}_{3} - (\boldsymbol{\phi}_{wb}^{b}\times)$. $\boldsymbol{\phi}_{wb}^{b}$ is the attitude error expressed in the $b$-frame.

According to equation (4) and (13), the differential equation of right error state in $w$-frame can be derived as (31).

$$\frac{d}{dt}\boldsymbol{\eta}_{wb}^{wR} = \left(\frac{d}{dt}\boldsymbol{\chi}_{wb}^{w}\right)\left(\tilde{\boldsymbol{\chi}}_{wb}^{w}\right)^{-1} - \boldsymbol{\eta}_{wb}^{wR}\left(\frac{d}{dt}\tilde{\boldsymbol{\chi}}_{wb}^{w}\right)\left(\tilde{\boldsymbol{\chi}}_{wb}^{w}\right)^{-1} \quad (31)$$

Expanding and simplifying the above equation using (1), (8) and (10), we get the differential equation of right error in matrix form in (32).

$$\frac{d}{dt}\boldsymbol{\eta}_{wb}^{wR} = \begin{bmatrix} -(\boldsymbol{\omega}_{ie}^w \times) & \boldsymbol{g}_{eb}^w & \mathbf{0}_{3\times 1} \\ \mathbf{0}_{1\times 3} & 0 & -1 \\ \mathbf{0}_{1\times 3} & 0 & 0 \end{bmatrix} \boldsymbol{\eta}_{wb}^{wR} - \boldsymbol{\eta}_{wb}^{wR} \begin{bmatrix} -(\boldsymbol{\omega}_{ie}^w \times) & \tilde{\boldsymbol{g}}_{eb}^w & \mathbf{0}_{3\times 1} \\ \mathbf{0}_{1\times 3} & 0 & -1 \\ \mathbf{0}_{1\times 3} & 0 & 0 \end{bmatrix}$$

$$-\boldsymbol{\eta}_{wb}^{wR} \tilde{\boldsymbol{\chi}}_{wb}^{w} \begin{bmatrix} (\delta\boldsymbol{\omega}_{ib}^b \times) & \delta\boldsymbol{f}_{ib}^b & \mathbf{0}_{3\times 1} \\ \mathbf{0}_{1\times 3} & 0 & 0 \\ \mathbf{0}_{1\times 3} & 0 & 0 \end{bmatrix} (\tilde{\boldsymbol{\chi}}_{wb}^{w})^{-1}$$

$$+ \left\{ \begin{bmatrix} -(\boldsymbol{\omega}_{ie}^w \times) & \mathbf{0}_{3\times 1} & \mathbf{0}_{3\times 1} \\ \mathbf{0}_{1\times 3} & 0 & 0 \\ \mathbf{0}_{1\times 3} & 0 & 0 \end{bmatrix} \boldsymbol{\eta}_{wb}^{wR} \\ -\boldsymbol{\eta}_{wb}^{wR} \begin{bmatrix} -(\boldsymbol{\omega}_{ie}^w \times) & \mathbf{0}_{3\times 1} & \mathbf{0}_{3\times 1} \\ \mathbf{0}_{1\times 3} & 0 & 0 \\ \mathbf{0}_{1\times 3} & 0 & 0 \end{bmatrix} \right\} \tilde{\boldsymbol{\chi}}_{wb}^{w} \begin{bmatrix} \mathbf{0}_{3\times 3} & \mathbf{0}_{3\times 1} & \mathbf{0}_{3\times 1} \\ \mathbf{0}_{1\times 3} & 1 & 0 \\ \mathbf{0}_{1\times 3} & 0 & -1 \end{bmatrix} (\tilde{\boldsymbol{\chi}}_{wb}^{w})^{-1}$$

(32)

Similarly, the differential equation of left error is given in (33).

$$\frac{d}{dt}\boldsymbol{\eta}_{wb}^{bL} = (\tilde{\boldsymbol{\chi}}_{wb}^{w})^{-1}\left(\frac{d}{dt}\boldsymbol{\chi}_{wb}^{w}\right) - (\tilde{\boldsymbol{\chi}}_{wb}^{w})^{-1}\left(\frac{d}{dt}\tilde{\boldsymbol{\chi}}_{wb}^{w}\right)\boldsymbol{\eta}_{wb}^{bL} \quad (33)$$

Similarly, we get the differential equation of left error in w-frame as shown in (34).

$$\frac{d}{dt}\boldsymbol{\eta}_{wb}^{bL} = \boldsymbol{\eta}_{wb}^{bL}\begin{bmatrix} (\boldsymbol{\omega}_{ib}^b \times) & \boldsymbol{f}_{ib}^b & \mathbf{0}_{3\times 1} \\ \mathbf{0}_{1\times 3} & 0 & 1 \\ \mathbf{0}_{1\times 3} & 0 & 0 \end{bmatrix} - \begin{bmatrix} (\tilde{\boldsymbol{\omega}}_{ib}^b \times) & \tilde{\boldsymbol{f}}_{ib}^b & \mathbf{0}_{3\times 1} \\ \mathbf{0}_{1\times 3} & 0 & 1 \\ \mathbf{0}_{1\times 3} & 0 & 0 \end{bmatrix}\boldsymbol{\eta}_{wb}^{bL}$$

$$-(\tilde{\boldsymbol{\chi}}_{wb}^{w})^{-1}\begin{bmatrix} \mathbf{0}_{3\times 3} & \delta\boldsymbol{g}_{wb}^e & \mathbf{0}_{3\times 1} \\ \mathbf{0}_{1\times 3} & 0 & -1 \\ \mathbf{0}_{1\times 3} & 0 & 0 \end{bmatrix}\boldsymbol{\chi}_{wb}^{w}$$

$$+(\tilde{\boldsymbol{\chi}}_{wb}^{w})^{-1}\begin{bmatrix} -(\boldsymbol{\omega}_{ie}^w \times) & \mathbf{0}_{3\times 1} & \mathbf{0}_{3\times 1} \\ \mathbf{0}_{1\times 3} & 0 & 0 \\ \mathbf{0}_{1\times 3} & 0 & 0 \end{bmatrix}\tilde{\boldsymbol{\chi}}_{wb}^{w} \left\{ \boldsymbol{\eta}_{wb}^{bL}\begin{bmatrix} \mathbf{0}_{3\times 3} & \mathbf{0}_{3\times 1} & \mathbf{0}_{3\times 1} \\ \mathbf{0}_{1\times 3} & 1 & 0 \\ \mathbf{0}_{1\times 3} & 0 & -1 \end{bmatrix} \\ - \begin{bmatrix} \mathbf{0}_{3\times 3} & \mathbf{0}_{3\times 1} & \mathbf{0}_{3\times 1} \\ \mathbf{0}_{1\times 3} & 1 & 0 \\ \mathbf{0}_{1\times 3} & 0 & -1 \end{bmatrix}\boldsymbol{\eta}_{wb}^{bL} \right\}$$

(34)

## III. DISCUSSION ON AUTONOMY

The propagation of system (error) state generally depends on the current state and the input, but there exists a class of systems that can be independent of the current state. This property makes the state propagation have a more concise form, and this property is also known as the autonomous property of (error) state propagation.

This section discusses the autonomous property of the navigation models under inertial frame, earth frame and world frame.

### A. Autonomous error propagation property

**Definition 1:** error state $e$ is told to have autonomous propagation property if there exist a function $g$ such that for any couple of solutions $\chi, \hat{\chi}$ of the dynamics, $e_t = e(\chi, \tilde{\chi})$ verifies:

$$\frac{d}{dt}e_t = g(e_t) \quad (35)$$

In this paper, it is referred to as the perfect autonomous error propagation property, which will be distinguished from the approximate autonomous error propagation property and the weak autonomous error propagation property discussed subsequently.

### B. Perfect, approximate and weak autonomy

If the two error forms, namely the left error $e_t = \boldsymbol{\eta}^L = \tilde{\boldsymbol{\chi}}^{-1}\boldsymbol{\chi}$ and the right error $e_t = \boldsymbol{\eta}^R = \boldsymbol{\chi}\tilde{\boldsymbol{\chi}}^{-1}$, can both satisfy the autonomous error propagation property, the system is said to have perfect autonomy.

**Theorem 1** (*perfect autonomy*): If $\chi$ and $\hat{\chi}$ can be expressed as derivatives with respect to time in the form (36).

$$\frac{d}{dt}\boldsymbol{\chi} = \boldsymbol{\chi}W_1 + W_2\boldsymbol{\chi}, \frac{d}{dt}\tilde{\boldsymbol{\chi}} = \tilde{\boldsymbol{\chi}}W_1 + W_2\tilde{\boldsymbol{\chi}} \quad (36)$$

For the right error and left error, the differential equation models exhibit perfect autonomous error propagation property, indicating that the error differential equations solely incorporate error terms and remain independent of $\chi, \hat{\chi}$. Here, the matrices $W_1$ and $W_2$ are independent of $\chi, \hat{\chi}$.

**Proof :** According to $\chi = \boldsymbol{\eta}^R \tilde{\chi}$ and $\chi = \tilde{\chi}\boldsymbol{\eta}^L$, we have:

$$\frac{d}{dt}\boldsymbol{\chi} = \frac{d}{dt}(\boldsymbol{\eta}^R \tilde{\chi}) = \left(\frac{d}{dt}\boldsymbol{\eta}^R\right)\tilde{\chi} + \boldsymbol{\eta}^R\left(\frac{d}{dt}\tilde{\chi}\right) \quad (37)$$

$$\frac{d}{dt}\boldsymbol{\chi} = \frac{d}{dt}(\tilde{\chi}\boldsymbol{\eta}^L) = \left(\frac{d}{dt}\tilde{\chi}\right)\boldsymbol{\eta}^L + \tilde{\chi}\left(\frac{d}{dt}\boldsymbol{\eta}^L\right) \quad (38)$$

Rearranging the terms, we get

$$\frac{d}{dt}\boldsymbol{\eta}^R = \left(\frac{d}{dt}\boldsymbol{\chi}\right)\tilde{\chi}^{-1} - \boldsymbol{\eta}^R\left(\frac{d}{dt}\tilde{\chi}\right)\tilde{\chi}^{-1} \quad (39)$$

$$\frac{d}{dt}\boldsymbol{\eta}^L = \tilde{\chi}^{-1}\left(\frac{d}{dt}\boldsymbol{\chi}\right) - \tilde{\chi}^{-1}\left(\frac{d}{dt}\tilde{\chi}\right)\boldsymbol{\eta}^L \quad (40)$$

Substituting (36) into (39)(40), we obtain (41)(42):

$$\frac{d}{dt}\boldsymbol{\eta}^R = (\boldsymbol{\chi}W_1 + W_2\boldsymbol{\chi})\tilde{\chi}^{-1} - \boldsymbol{\chi}\tilde{\chi}^{-1}(\tilde{\chi}W_1 + W_2\tilde{\chi})\tilde{\chi}^{-1} \\ = W_2\boldsymbol{\eta}^R - \boldsymbol{\eta}^R W_2 \quad (41)$$

$$\frac{d}{dt}\boldsymbol{\eta}^L = \tilde{\chi}^{-1}(\boldsymbol{\chi}W_1 + W_2\boldsymbol{\chi}) - \tilde{\chi}^{-1}(\tilde{\chi}W_1 + W_2\tilde{\chi})\tilde{\chi}^{-1}\boldsymbol{\chi} \\ = \boldsymbol{\eta}^L W_1 - W_1\boldsymbol{\eta}^L \quad (42)$$

The right error differential equation only contains the right error term $\boldsymbol{\eta}^R$, and the left error differential equation only contains the left error term $\boldsymbol{\eta}^L$. Both are independent of $\chi$, $\tilde{\chi}$ and, that is, the right error and left error differential equation models both have perfect autonomous error propagation.

**Corollary 1** (*approximate autonomy*): If $\chi$ and $\hat{\chi}$ can be expressed as derivatives in the form (43).

$$\frac{d}{dt}\chi = \chi W_1 + W_2\chi, \frac{d}{dt}\tilde{\chi} = \tilde{\chi}\tilde{W}_1 + \tilde{W}_2\tilde{\chi} \quad (43)$$

Then the left and right error differential equation can be derived as (44) and (45).

$$\begin{aligned}\frac{d}{dt}\eta^R &= \left(\frac{d}{dt}\chi\right)\tilde{\chi}^{-1} - \eta^R\left(\frac{d}{dt}\tilde{\chi}\right)\tilde{\chi}^{-1} \\ &= (\chi W_1 + W_2\chi)\tilde{\chi}^{-1} - \chi\tilde{\chi}^{-1}\left(\tilde{\chi}\tilde{W}_1 + \tilde{W}_2\tilde{\chi}\right)\tilde{\chi}^{-1} \\ &= W_2\chi\tilde{\chi}^{-1} - \chi\tilde{\chi}^{-1}\tilde{W}_2 - \chi\tilde{\chi}^{-1}\tilde{\chi}\left(\tilde{W}_1 - W_1\right)\tilde{\chi}^{-1} \\ &= W_2\eta^R - \eta^R\tilde{W}_2 - \eta^R\left(\tilde{\chi}\delta W_1\tilde{\chi}^{-1}\right)\end{aligned} \quad (44)$$

$$\begin{aligned}\frac{d}{dt}\eta^L &= \tilde{\chi}^{-1}\left(\frac{d}{dt}\chi\right) - \tilde{\chi}^{-1}\left(\frac{d}{dt}\tilde{\chi}\right)\eta^L \\ &= \tilde{\chi}^{-1}(\chi W_1 + W_2\chi) - \tilde{\chi}^{-1}\left(\tilde{\chi}\tilde{W}_1 + \tilde{W}_2\tilde{\chi}\right)\eta^L \\ &= \tilde{\chi}^{-1}\chi W_1 - \tilde{W}_1\tilde{\chi}^{-1}\chi - \tilde{\chi}^{-1}\left(\tilde{W}_2 - W_2\right)\tilde{\chi}\eta^L \\ &= \eta^L W_1 - \tilde{W}_1\eta^L - \left(\tilde{\chi}^{-1}\delta W_2\tilde{\chi}\right)\eta^L\end{aligned} \quad (45)$$

Due to the existence of $\delta W_1$, $\delta W_2$, the differential equation contains matrix Lie group elements with errors $\tilde{\chi}$, and thus none of them possess perfect autonomous error propagation property. This system is said to has approximate autonomous error propagation property.

**Corollary 2** (*weak autonomy*): If $\chi$ and $\hat{\chi}$ can be expressed as derivatives with respect to time in the form (46).

$$\frac{d}{dt}\chi = \chi W_1 + W_2\chi, \frac{d}{dt}\tilde{\chi} = \tilde{\chi}\tilde{W}_1 + \tilde{W}_2\tilde{\chi} + W_3\tilde{\chi}W_4 \quad (46)$$

Then the left and right error differential equation can be derived as (47) and (48).

$$\begin{aligned}\frac{d}{dt}\eta^R &= \left(\frac{d}{dt}\chi\right)\tilde{\chi}^{-1} - \eta^R\left(\frac{d}{dt}\tilde{\chi}\right)\tilde{\chi}^{-1} \\ &= (\chi W_1 + W_2\chi + W_3\chi W_4)\tilde{\chi}^{-1} - \eta^R\left(\tilde{\chi}\tilde{W}_1 + \tilde{W}_2\tilde{\chi} + W_3\tilde{\chi}W_4\right)\tilde{\chi}^{-1} \\ &= W_2\eta^R - \eta^R\tilde{W}_2 - \eta^R\tilde{\chi}\delta W_1\tilde{\chi}^{-1} + \left(W_3\eta^R - \eta^R W_3\right)\tilde{\chi}W_4\tilde{\chi}^{-1}\end{aligned} \quad (47)$$

$$\begin{aligned}\frac{d}{dt}\eta^L &= \tilde{\chi}^{-1}\left(\frac{d}{dt}\chi\right) - \tilde{\chi}^{-1}\left(\frac{d}{dt}\tilde{\chi}\right)\eta^L \\ &= \tilde{\chi}^{-1}(\chi W_1 + W_2\chi + W_3\chi W_4) - \tilde{\chi}^{-1}\left(\tilde{\chi}\tilde{W}_1 + \tilde{W}_2\tilde{\chi} + W_3\tilde{\chi}W_4\right)\eta^L \\ &= \eta^L W_1 - \tilde{W}_1\eta^L - \left(\tilde{\chi}^{-1}\delta W_2\tilde{\chi}\right)\eta^L + \tilde{\chi}^{-1}W_3\tilde{\chi}\left(\eta^L W_4 - W_4\eta^L\right)\end{aligned} \quad (48)$$

Due to the existence of $\delta W_1$, $\delta W_2$, $W_3$ and $W_4$. The differential equation contains group elements with error $\tilde{\chi}$, and thus none of them possess perfect autonomous error propagation property. Due to the existence of $W_3$ and $W_4$, the autonomous property of Corollary 2 is said to be weaker than that in Corollary 1.

For the inertial frame navigation model, $W_2$ only contains gravitational terms without the inclusion of the $W_3$ and $W_4$ terms. When gyroscope and accelerometer bias are neglected, along with gravitational field errors caused by position errors, the case conforms to perfect autonomy (Theorem 1). Conversely, when considering gyroscope and accelerometer errors alongside gravitational field errors induced by position errors, the case corresponds to approximate autonomy (Corollary 1). For earth frame and world frame navigation error models, the models incorporate the $W_3$ and $W_4$ terms, which are associated with Coriolis force. When accounting for gyroscope and accelerometer bias, as well as gravitational field errors arising from position errors, the situation conforms to weak autonomy (Corollary 2).

If gyroscope bias, accelerometer bias, and gravitational field error due to position errors is negligible, and Coriolis force is minimal (e.g. at low speed), the influence of the related $W_3$ and $W_4$ terms remains insignificant. Consequently, the $SE_2(3)$ group based navigation model could exhibit favorable approximate autonomous error property in most applications. Nevertheless, improving autonomy is preferred if possible.

Therefore, in this paper, a new construction method is proposed for the earth frame and world frame navigation error model respectively to eliminate the effects of $W_3$ and $W_4$ terms, and thereby further enhancing the autonomous properties of the model, which is shown is the next section.

## IV. IMPROVE AUTONOMY OF NON-INERTIAL NAVIGATION MODEL

In the navigation model of non-inertial frames, the main reason for the loss of autonomy is that the differential equation of velocity includes Coriolis force caused by the earth's rotation. Therefore, a construction method is proposed, where the inertial frame velocity is used to replace the non-inertial frame velocity. The proposed $SE_2(3)$ group element is given as in (49).

$$\begin{aligned}\chi_{ieb}^e &= \begin{bmatrix} C_b^e & v_{ib}^e - (\omega_{ie}^e \times)r_{ib(0)}^e & r_{ib}^e - r_{ib(0)}^e \\ 0_{1\times3} & 1 & 0 \\ 0_{1\times3} & 0 & 1 \end{bmatrix} \\ &= \begin{bmatrix} C_b^e & v_{ib}^e - (\omega_{ie}^e \times)r_{eb(0)}^e & r_{eb}^e - r_{eb(0)}^e \\ 0_{1\times3} & 1 & 0 \\ 0_{1\times3} & 0 & 1 \end{bmatrix} \in SE_2(3)\end{aligned} \quad (49)$$

where $v_{ib}^e$ minus $(\omega_{ie}^e \times)r_{ib(0)}^e$ to ensure the initial velocity is a small element.

Its differential equation is derived as in (50).

$$\frac{d}{dt}\left(\boldsymbol{\chi}_{ieb}^{e}\right)$$

$$=\begin{bmatrix} \boldsymbol{C}_{b}^{e} & \boldsymbol{v}_{ib}^{e}-\left(\boldsymbol{\omega}_{ie}^{e}\times\right)\boldsymbol{r}_{eb(0)}^{e} & \boldsymbol{r}_{eb}^{e}-\boldsymbol{r}_{eb(0)}^{e} \\ \boldsymbol{0}_{1\times 3} & 1 & 0 \\ \boldsymbol{0}_{1\times 3} & 0 & 1 \end{bmatrix}\begin{bmatrix} \left(\boldsymbol{\omega}_{ib}^{b}\times\right) & \boldsymbol{f}_{ib}^{b} & \boldsymbol{0}_{3\times 1} \\ \boldsymbol{0}_{1\times 3} & 0 & 1 \\ \boldsymbol{0}_{1\times 3} & 0 & 0 \end{bmatrix}$$

$$+\begin{bmatrix} -\left(\boldsymbol{\omega}_{ie}^{e}\times\right) & \boldsymbol{\gamma}_{ib}^{e}-\left(\boldsymbol{\omega}_{ie}^{e}\times\right)^{2}\boldsymbol{r}_{eb(0)}^{e} & \boldsymbol{0}_{3\times 1} \\ \boldsymbol{0}_{1\times 3} & 0 & -1 \\ \boldsymbol{0}_{1\times 3} & 0 & 0 \end{bmatrix} \quad (50)$$

$$\begin{bmatrix} \boldsymbol{C}_{b}^{e} & \boldsymbol{v}_{ib}^{e}-\left(\boldsymbol{\omega}_{ie}^{e}\times\right)\boldsymbol{r}_{eb(0)}^{e} & \boldsymbol{r}_{eb}^{e}-\boldsymbol{r}_{eb(0)}^{e} \\ \boldsymbol{0}_{1\times 3} & 1 & 0 \\ \boldsymbol{0}_{1\times 3} & 0 & 1 \end{bmatrix}$$

Define the right error and left error as in (51) and (52).

$$\boldsymbol{\eta}_{ieb}^{eR}=\boldsymbol{\chi}_{ieb}^{e}\left(\tilde{\boldsymbol{\chi}}_{ieb}^{e}\right)^{-1}$$

$$=\begin{bmatrix} \boldsymbol{C}_{b}^{e} & \boldsymbol{v}_{ib}^{e}-\left(\boldsymbol{\omega}_{ie}^{e}\times\right)\boldsymbol{r}_{eb(0)}^{e} & \boldsymbol{r}_{eb}^{e}-\boldsymbol{r}_{eb(0)}^{e} \\ \boldsymbol{0}_{1\times 3} & 1 & 0 \\ \boldsymbol{0}_{1\times 3} & 0 & 1 \end{bmatrix}\begin{bmatrix} \tilde{\boldsymbol{C}}_{b}^{e} & \tilde{\boldsymbol{v}}_{ib}^{e}-\left(\boldsymbol{\omega}_{ie}^{e}\times\right)\boldsymbol{r}_{eb(0)}^{e} & \tilde{\boldsymbol{r}}_{eb}^{e}-\boldsymbol{r}_{eb(0)}^{e} \\ \boldsymbol{0}_{1\times 3} & 1 & 0 \\ \boldsymbol{0}_{1\times 3} & 0 & 1 \end{bmatrix}^{-1}$$

$$\approx\begin{bmatrix} \exp(\boldsymbol{\phi}_{eb}^{e}) & -\delta\boldsymbol{v}_{ib}^{e}+\left[\tilde{\boldsymbol{v}}_{ib}^{e}-\left(\boldsymbol{\omega}_{ie}^{e}\times\right)\boldsymbol{r}_{eb(0)}^{e}\right]\times\boldsymbol{\phi}_{eb}^{e} & -\delta\boldsymbol{r}_{eb}^{e}+\left(\tilde{\boldsymbol{r}}_{eb}^{e}-\boldsymbol{r}_{eb(0)}^{e}\right)\times\boldsymbol{\phi}_{eb}^{e} \\ \boldsymbol{0}_{1\times 3} & 1 & 0 \\ \boldsymbol{0}_{1\times 3} & 0 & 1 \end{bmatrix} \quad (51)$$

$$\boldsymbol{\eta}_{ieb}^{bL}=\left(\tilde{\boldsymbol{\chi}}_{ieb}^{e}\right)^{-1}\boldsymbol{\chi}_{ieb}^{e}$$

$$=\begin{bmatrix} \tilde{\boldsymbol{C}}_{b}^{e} & \tilde{\boldsymbol{v}}_{ib}^{e}-\left(\boldsymbol{\omega}_{ie}^{e}\times\right)\boldsymbol{r}_{eb(0)}^{e} & \tilde{\boldsymbol{r}}_{eb}^{e}-\boldsymbol{r}_{eb(0)}^{e} \\ \boldsymbol{0}_{1\times 3} & 1 & 0 \\ \boldsymbol{0}_{1\times 3} & 0 & 1 \end{bmatrix}^{-1}\begin{bmatrix} \boldsymbol{C}_{b}^{e} & \boldsymbol{v}_{ib}^{e}-\left(\boldsymbol{\omega}_{ie}^{e}\times\right)\boldsymbol{r}_{eb(0)}^{e} & \boldsymbol{r}_{eb}^{e}-\boldsymbol{r}_{eb(0)}^{e} \\ \boldsymbol{0}_{1\times 3} & 1 & 0 \\ \boldsymbol{0}_{1\times 3} & 0 & 1 \end{bmatrix}$$

$$=\begin{bmatrix} \boldsymbol{C}_{b'}^{b'} & -\boldsymbol{C}_{b}^{b'}\delta\boldsymbol{v}_{ib}^{b} & -\boldsymbol{C}_{b}^{b'}\delta\boldsymbol{r}_{eb}^{b} \\ \boldsymbol{0}_{1\times 3} & 1 & 0 \\ \boldsymbol{0}_{1\times 3} & 0 & 1 \end{bmatrix} \quad (52)$$

$$=\begin{bmatrix} \exp(-\boldsymbol{\phi}_{eb}^{b}\times) & -\exp(-\boldsymbol{\phi}_{eb}^{b}\times)\delta\boldsymbol{v}_{ib}^{b} & -\exp(-\boldsymbol{\phi}_{eb}^{b}\times)\delta\boldsymbol{r}_{ib}^{b} \\ \boldsymbol{0}_{1\times 3} & 1 & 0 \\ \boldsymbol{0}_{1\times 3} & 0 & 1 \end{bmatrix}$$

$$\approx\begin{bmatrix} \exp(-\boldsymbol{\phi}_{eb}^{b}\times) & -\delta\boldsymbol{v}_{ib}^{b} & -\delta\boldsymbol{r}_{eb}^{b} \\ \boldsymbol{0}_{1\times 3} & 1 & 0 \\ \boldsymbol{0}_{1\times 3} & 0 & 1 \end{bmatrix}$$

The corresponding right error differential equation is given in (53).

$$\frac{d}{dt}\boldsymbol{\eta}_{ieb}^{eR}=\left(\frac{d}{dt}\boldsymbol{\chi}_{ieb}^{e}\right)\left(\tilde{\boldsymbol{\chi}}_{ieb}^{e}\right)^{-1}-\boldsymbol{\eta}_{ieb}^{eR}\left(\frac{d}{dt}\tilde{\boldsymbol{\chi}}_{ieb}^{e}\right)\left(\tilde{\boldsymbol{\chi}}_{ieb}^{e}\right)^{-1}$$

$$=\begin{bmatrix} -\left(\boldsymbol{\omega}_{ie}^{e}\times\right) & \boldsymbol{\gamma}_{ib}^{e}-\left(\boldsymbol{\omega}_{ie}^{e}\times\right)^{2}\boldsymbol{r}_{eb(0)}^{e} & \boldsymbol{0}_{3\times 1} \\ \boldsymbol{0}_{1\times 3} & 0 & -1 \\ \boldsymbol{0}_{1\times 3} & 0 & 0 \end{bmatrix}\boldsymbol{\eta}_{ieb}^{eR}$$

$$-\boldsymbol{\eta}_{ieb}^{eR}\begin{bmatrix} -\left(\boldsymbol{\omega}_{ie}^{e}\times\right) & \boldsymbol{\gamma}_{ib}^{e}-\left(\boldsymbol{\omega}_{ie}^{e}\times\right)^{2}\boldsymbol{r}_{eb(0)}^{e} & \boldsymbol{0}_{3\times 1} \\ \boldsymbol{0}_{1\times 3} & 0 & -1 \\ \boldsymbol{0}_{1\times 3} & 0 & 0 \end{bmatrix} \quad (53)$$

$$-\boldsymbol{\eta}_{ieb}^{eR}\tilde{\boldsymbol{\chi}}_{ieb}^{e}\left(\begin{bmatrix} \left(\delta\boldsymbol{\omega}_{ib}^{b}\times\right) & \delta\boldsymbol{f}_{ib}^{b} & \boldsymbol{0}_{3\times 1} \\ \boldsymbol{0}_{1\times 3} & 0 & 0 \\ \boldsymbol{0}_{1\times 3} & 0 & 0 \end{bmatrix}\left(\tilde{\boldsymbol{\chi}}_{ieb}^{e}\right)^{-1}\right)$$

Similarly, the left error differential equation is given in (54).

$$\frac{d}{dt}\boldsymbol{\eta}_{ieb}^{bL}=\left(\tilde{\boldsymbol{\chi}}_{ieb}^{e}\right)^{-1}\left(\frac{d}{dt}\boldsymbol{\chi}_{ieb}^{e}\right)-\left(\tilde{\boldsymbol{\chi}}_{ieb}^{e}\right)^{-1}\left(\frac{d}{dt}\tilde{\boldsymbol{\chi}}_{ieb}^{e}\right)\boldsymbol{\eta}_{ieb}^{bL}$$

$$=\boldsymbol{\eta}_{ieb}^{bL}\begin{bmatrix} \left(\boldsymbol{\omega}_{ib}^{b}\times\right) & \boldsymbol{f}_{ib}^{b} & \boldsymbol{0}_{3\times 1} \\ \boldsymbol{0}_{1\times 3} & 0 & 1 \\ \boldsymbol{0}_{1\times 3} & 0 & 0 \end{bmatrix}-\begin{bmatrix} \left(\tilde{\boldsymbol{\omega}}_{ib}^{b}\times\right) & \tilde{\boldsymbol{f}}_{ib}^{b} & \boldsymbol{0}_{3\times 1} \\ \boldsymbol{0}_{1\times 3} & 0 & 1 \\ \boldsymbol{0}_{1\times 3} & 0 & 0 \end{bmatrix}\boldsymbol{\eta}_{ieb}^{bL} \quad (54)$$

$$+\left(\tilde{\boldsymbol{\chi}}_{ieb}^{e}\right)^{-1}\begin{bmatrix} \boldsymbol{0}_{3\times 3} & \delta\boldsymbol{\gamma}_{ib}^{e} & \boldsymbol{0}_{3\times 1} \\ \boldsymbol{0}_{1\times 3} & 0 & 0 \\ \boldsymbol{0}_{1\times 3} & 0 & 0 \end{bmatrix}\tilde{\boldsymbol{\chi}}_{ieb}^{e}\boldsymbol{\eta}_{ieb}^{bL}$$

Comparing (53) (54) with (27) (28), it can be found new model eliminates the terms of $W_3$ and $W_4$.

### A. General Formula of the proposed navigation model

Consider two projection frames: the earth frame and the world frame. The effects of gyroscope and accelerometer errors are still accounted for, and an inertial navigation error model with approximate autonomous property satisfying Corollary 1 is analyzed. To simplify the notation, the corresponding terms of Corollary 1 under different frame, namely $W_1, W_2, \tilde{W}_1, \tilde{W}_2$ and $\delta W_1, \delta W_2$ are not distinguished symbolically.

For $\Delta \boldsymbol{v}_{ib(0)}^{\gamma}, \gamma=i,e,w$, let $\Delta \boldsymbol{v}_{ib(0)}^{i}=\boldsymbol{0}_{3\times 1}$, $\Delta \boldsymbol{v}_{ib(0)}^{e}=\left(\boldsymbol{\omega}_{ie}^{e}\times\right)\boldsymbol{r}_{eb(0)}^{e}$, $\Delta \boldsymbol{v}_{ib(0)}^{w}=\left(\boldsymbol{\omega}_{ie}^{w}\times\right)\left(\boldsymbol{r}_{ew}^{w}+\boldsymbol{r}_{wb(0)}^{w}\right)$, the general form of the proposed right error of the $SE_2(3)$ group can be expressed as in (55).

$$\boldsymbol{\eta}_{i\gamma b}^{\gamma R}=\boldsymbol{\chi}_{i\gamma b}^{\gamma}\left(\tilde{\boldsymbol{\chi}}_{i\gamma b}^{\gamma}\right)^{-1}$$

$$=\begin{bmatrix} \boldsymbol{C}_{b}^{\gamma} & \boldsymbol{v}_{ib}^{\gamma}-\Delta\boldsymbol{v}_{ib(0)}^{\gamma} & \boldsymbol{r}_{\gamma b}^{\gamma}-\boldsymbol{r}_{\gamma b(0)}^{\gamma} \\ \boldsymbol{0}_{1\times 3} & 1 & 0 \\ \boldsymbol{0}_{1\times 3} & 0 & 1 \end{bmatrix}\begin{bmatrix} \tilde{\boldsymbol{C}}_{b}^{\gamma} & \tilde{\boldsymbol{v}}_{ib}^{\gamma}-\Delta\boldsymbol{v}_{ib(0)}^{\gamma} & \tilde{\boldsymbol{r}}_{\gamma b}^{\gamma}-\boldsymbol{r}_{\gamma b(0)}^{\gamma} \\ \boldsymbol{0}_{1\times 3} & 1 & 0 \\ \boldsymbol{0}_{1\times 3} & 0 & 1 \end{bmatrix}^{-1}$$

$$\approx\begin{bmatrix} \exp(\boldsymbol{\phi}_{\gamma b}^{\gamma}\times) & -\delta\boldsymbol{v}_{ib}^{\gamma}+\left(\tilde{\boldsymbol{v}}_{ib}^{\gamma}-\Delta\boldsymbol{v}_{ib(0)}^{\gamma}\right)\times\boldsymbol{\phi}_{\gamma b}^{\gamma} & -\delta\boldsymbol{r}_{\gamma b}^{\gamma}+\left(\tilde{\boldsymbol{r}}_{\gamma b}^{\gamma}-\boldsymbol{r}_{\gamma b(0)}^{\gamma}\right)\times\boldsymbol{\phi}_{\gamma b}^{\gamma} \\ \boldsymbol{0}_{1\times 3} & 1 & 0 \\ \boldsymbol{0}_{1\times 3} & 0 & 1 \end{bmatrix}$$

$$\triangleq\begin{bmatrix} \exp(\boldsymbol{\phi}_{\gamma b}^{\gamma}\times) & \boldsymbol{J\rho}_{vib}^{\gamma R} & \boldsymbol{J\rho}_{r\gamma b}^{\gamma R} \\ \boldsymbol{0}_{1\times 3} & 1 & 0 \\ \boldsymbol{0}_{1\times 3} & 0 & 1 \end{bmatrix}$$

(55)

Note that the subscripts for $\boldsymbol{\phi}_{\gamma b}^{\gamma}$, $\boldsymbol{J\rho}_{vib}^{\gamma R}$ and $\boldsymbol{J\rho}_{r\gamma b}^{\gamma R}$ are explicitly specified here, where $\gamma$ denotes the projected frame and also serving as the reference frame for position, while the inertial frame functions as the reference frame for velocity. Let $R$ represent the right error and $L$ denote the left error.

The attitude, velocity, and position error states corresponding to the right error are represented as (56).

$$\xi_{i\gamma b}^{\gamma R} = \begin{bmatrix} \phi_{\gamma b}^{\gamma} \\ J\rho_{vib}^{\gamma R} \\ J\rho_{r\gamma b}^{\gamma R} \end{bmatrix} = \begin{bmatrix} \phi_{\gamma b}^{\gamma} \\ -\delta v_{ib}^{\gamma} + \left[I_3 - \exp(\phi_{\gamma b}^{\gamma} \times)\right]\left(\tilde{v}_{ib}^{\gamma} - \Delta v_{ib(0)}^{\gamma}\right) \\ -\delta r_{\gamma b}^{\gamma} + \left[I_3 - \exp(\phi_{\gamma b}^{\gamma} \times)\right]\left(\tilde{r}_{\gamma b}^{\gamma} - r_{\gamma b(0)}^{\gamma}\right) \end{bmatrix}$$

$$\approx \begin{bmatrix} \phi_{\gamma b}^{\gamma} \\ -\delta v_{ib}^{\gamma} + \left(\tilde{v}_{ib}^{\gamma} - \Delta v_{ib(0)}^{\gamma}\right) \times \phi_{\gamma b}^{\gamma} \\ -\delta r_{\gamma b}^{\gamma} + \left(\tilde{r}_{\gamma b}^{\gamma} - r_{\gamma b(0)}^{\gamma}\right) \times \phi_{\gamma b}^{\gamma} \end{bmatrix} \quad (56)$$

$$\gamma = i, e, w$$

The general form of the proposed right error of the $SE_2(3)$ group can be expressed as in (57).

$$\eta_{i\gamma b}^{bL} = \left(\tilde{\chi}_{i\gamma b}^{\gamma}\right)^{-1} \chi_{i\gamma b}^{\gamma}$$

$$= \begin{bmatrix} \tilde{C}_b^{\gamma} & \tilde{v}_{ib}^{\gamma} - \Delta v_{ib(0)}^{\gamma} & \tilde{r}_{\gamma b}^{\gamma} - r_{\gamma b(0)}^{\gamma} \\ 0_{1\times 3} & 1 & 0 \\ 0_{1\times 3} & 0 & 1 \end{bmatrix}^{-1} \begin{bmatrix} C_b^{\gamma} & v_{ib}^{\gamma} - \Delta v_{ib(0)}^{\gamma} & r_{\gamma b}^{\gamma} - r_{\gamma b(0)}^{\gamma} \\ 0_{1\times 3} & 1 & 0 \\ 0_{1\times 3} & 0 & 1 \end{bmatrix}$$

$$= \begin{bmatrix} \exp(-\phi_{\gamma b}^b \times) & -\exp(-\phi_{\gamma b}^b \times)\delta v_{ib}^b & -\exp(-\phi_{\gamma b}^b \times)\delta r_{\gamma b}^b \\ 0_{1\times 3} & 1 & 0 \\ 0_{1\times 3} & 0 & 1 \end{bmatrix}$$

$$\approx \begin{bmatrix} \exp(-\phi_{\gamma b}^b \times) & -\delta v_{ib}^b & -\delta r_{\gamma b}^b \\ 0_{1\times 3} & 1 & 0 \\ 0_{1\times 3} & 0 & 1 \end{bmatrix} \triangleq \begin{bmatrix} \exp(-\phi_{\gamma b}^b \times) & J\rho_{vib}^{bL} & J\rho_{r\gamma b}^{bL} \\ 0_{1\times 3} & 1 & 0 \\ 0_{1\times 3} & 0 & 1 \end{bmatrix}$$

(57)

The attitude, velocity, and position error states corresponding to the left error are represented as (58).

$$\xi_{i\gamma b}^{bL} = \begin{bmatrix} \phi_{\gamma b}^b \\ J\rho_{vi\gamma b}^{bL} \\ J\rho_{r\gamma b}^{\gamma L} \end{bmatrix} = \begin{bmatrix} \phi_{\gamma b}^b \\ -\exp(-\phi_{\gamma b}^b \times)\delta v_{ib}^b \\ -\exp(-\phi_{\gamma b}^b \times)\delta r_{\gamma b}^b \end{bmatrix} \approx \begin{bmatrix} \phi_{\gamma b}^b \\ -\delta v_{ib}^b \\ -\delta r_{\gamma b}^b \end{bmatrix}, \quad \gamma = i, e, w \quad (58)$$

The true navigation state of $SE_2(3)$ group can be recovered and calculated by (59).

$$\chi_{i\gamma b}^{\gamma} = \eta_{i\gamma b}^{\gamma R} \tilde{\chi}_{i\gamma b}^{\gamma} = \begin{bmatrix} \exp(\phi_{\gamma b}^{\gamma} \times) & J\rho_{vi\gamma b}^{\gamma R} & J\rho_{r\gamma b}^{\gamma R} \\ 0_{1\times 3} & 1 & 0 \\ 0_{1\times 3} & 0 & 1 \end{bmatrix} \tilde{\chi}_{\beta b}^{\gamma}$$

$$\chi_{i\gamma b}^{\gamma} = \tilde{\chi}_{i\gamma b}^{\gamma} \eta_{i\gamma b}^{bL} = \tilde{\chi}_{i\gamma b}^{\gamma} \begin{bmatrix} \exp(-\phi_{\gamma b}^b \times) & J\rho_{vi\gamma b}^{bL} & J\rho_{r\gamma b}^{bL} \\ 0_{1\times 3} & 1 & 0 \\ 0_{1\times 3} & 0 & 1 \end{bmatrix}$$

(59)

where

$$C_b^{\gamma} = \exp(\phi_{\gamma b}^{\gamma} \times)\tilde{C}_b^{\gamma} = \tilde{C}_b^{\gamma} \exp(-\phi_{\gamma b}^b \times)$$

$$\begin{cases} \exp(\phi_{\gamma b}^{\gamma} \times) \approx I_3 + (\phi_{\gamma b}^{\gamma} \times) \\ \exp(-\phi_{\gamma b}^b \times) \approx I_3 - (\phi_{\gamma b}^b \times) \end{cases} \quad (60)$$

### B. Derivation of the Differential Equations

#### 1) $i$-frame navigation error state and its derivative

The right error form and left error form of inertial frame is given in (61) and (62).

$$\gamma = i \Rightarrow \xi_{i\gamma b}^{\gamma R} = \xi_{ib}^{iR} = \begin{bmatrix} \phi_{ib}^i \\ J\rho_{vib}^{iR} \\ J\rho_{rib}^{iR} \end{bmatrix}$$

$$= \begin{bmatrix} \phi_{ib}^i \\ -\delta v_{ib}^i + \left[I_3 - \exp(\phi_{ib}^i \times)\right]\tilde{v}_{ib}^i \\ -\delta r_{ib}^i + \left[I_3 - \exp(\phi_{ib}^i \times)\right]\left(\tilde{r}_{ib}^i - r_{ib(0)}^i\right) \end{bmatrix} \approx \begin{bmatrix} \phi_{ib}^i \\ -\delta v_{ib}^i + \left(\tilde{v}_{ib}^i \times\right)\phi_{ib}^i \\ -\delta r_{ib}^i + \left[\left(\tilde{r}_{ib}^i - r_{ib(0)}^i\right) \times\right]\phi_{ib}^i \end{bmatrix}$$

(61)

where $\xi_{iib}^{iR}$ is simplified as $\xi_{ib}^{iR}$ and $J\rho_{viib}^{iR}$ is simplified as $J\rho_{vib}^{iR}$.

$$\gamma = i \Rightarrow \xi_{i\gamma b}^{bL} = \xi_{ib}^{bL} = \begin{bmatrix} \phi_{ib}^b \\ J\rho_{vib}^{bL} \\ J\rho_{rib}^{bL} \end{bmatrix} \approx \begin{bmatrix} \phi_{ib}^b \\ -\delta v_{ib}^b \\ -\delta r_{ib}^b \end{bmatrix} \quad (62)$$

where $\xi_{iib}^{iL}$ is simplified as $\xi_{ib}^{bL}$ and $J\rho_{viib}^{iL}$ is simplified as $J\rho_{vib}^{iL}$.

The differential equation of attitude error in the inertial frame is:

$$\dot{\phi}_{ib}^i = -\tilde{C}_b^i \delta \omega_{ib}^b \approx -C_b^i \delta \omega_{ib}^b \quad (63)$$

The differential equation of attitude error in the body frame is:

$$\dot{\phi}_{ib}^b = -\left(\tilde{\omega}_{ib}^b \times\right)\phi_{ib}^b + \delta \omega_{ib}^b \quad (64)$$

According to Corollary 1 and (55), the differential equation of right error can be derived as in (65).

$$\frac{d}{dt}\eta_{ib}^{iR} = W_2 \eta_{ib}^{iR} - \eta_{ib}^{iR}\tilde{W}_2 - \eta_{ib}^{iR}\left(\tilde{\chi}_{ib}^i \delta W_1 \left(\tilde{\chi}_{ib}^i\right)^{-1}\right)$$

$$= \begin{bmatrix} 0_{3\times 3} & \gamma_{ib}^i & 0_{3\times 1} \\ 0_{1\times 3} & 0 & -1 \\ 0_{1\times 3} & 0 & 0 \end{bmatrix}\eta_{ib}^{iR} - \eta_{ib}^{iR}\begin{bmatrix} 0_{3\times 3} & \tilde{\gamma}_{ib}^i & 0_{3\times 1} \\ 0_{1\times 3} & 0 & -1 \\ 0_{1\times 3} & 0 & 0 \end{bmatrix}$$

$$-\eta_{ib}^{iR}\tilde{\chi}_{ib}^i\begin{bmatrix} \left(\delta\omega_{ib}^b \times\right) & \delta f_{ib}^b & 0_{3\times 1} \\ 0_{1\times 3} & 0 & 0 \\ 0_{1\times 3} & 0 & 0 \end{bmatrix}\tilde{\chi}_{ib}^{i-1}$$

(65)

where

$$\eta_{ib}^{iR} = \begin{bmatrix} \exp(\phi_{ib}^i \times) & J\rho_{vib}^{iR} & J\rho_{rib}^{iR} \\ 0_{1\times 3} & 1 & 0 \\ 0_{1\times 3} & 0 & 1 \end{bmatrix}$$

$$\tilde{\chi}_{ib}^i = \begin{bmatrix} \tilde{C}_b^i & \tilde{v}_{ib}^i & \tilde{r}_{ib}^i - r_{ib(0)}^i \\ 0_{1\times 3} & 1 & 0 \\ 0_{1\times 3} & 0 & 1 \end{bmatrix}$$

(66)

Rewrite (65) to a vector form as shown in (67),(68).

$$J\dot{\rho}_{vib}^{iR} = \gamma_{ib}^i - \exp(\phi_{ib}^i \times)\tilde{\gamma}_{ib}^i + \exp(\phi_{ib}^i \times)\tilde{C}_b^i(\delta\omega_{ib}^b \times)\tilde{C}_i^b \tilde{v}_{ib}^i$$
$$- \exp(\phi_{ib}^i \times)\tilde{C}_b^i \delta f_{ib}^b$$
$$= -\delta\gamma_{ib}^i + \left[I_3 - \exp(\phi_{ib}^i \times)\right]\tilde{\gamma}_{ib}^i$$
$$- \exp(\phi_{ib}^i \times)(\tilde{v}_{ib}^i \times)\tilde{C}_b^i \delta\omega_{ib}^b - \exp(\phi_{ib}^i \times)\tilde{C}_b^i \delta f_{ib}^b \quad (67)$$

$$J\dot{\rho}_{rib}^{iR} = J\dot{\rho}_{vib}^{iR} + \exp(\phi_{ib}^i \times)\tilde{C}_b^i(\delta\omega_{ib}^b \times)\tilde{C}_i^b(\tilde{r}_{ib}^i - r_{ib(0)}^i)$$
$$= J\dot{\rho}_{vib}^{iR} - \exp(\phi_{ib}^i \times)\left[(\tilde{r}_{ib}^i - r_{ib(0)}^i) \times\right]\tilde{C}_b^i \delta\omega_{ib}^b \quad (68)$$

Performing a first-order approximation yields:

$$J\dot{\rho}_{vib}^{iR} \approx (\tilde{\gamma}_{ib}^i \times)\phi_{ib}^i - \delta\gamma_{ib}^i - (\tilde{v}_{ib}^i \times)\tilde{C}_b^i \delta\omega_{ib}^b - \tilde{C}_b^i \delta f_{ib}^b \quad (69)$$

$$J\dot{\rho}_{rib}^{iR} \approx J\dot{\rho}_{vib}^{iR} - \left[(\tilde{r}_{ib}^i - r_{ib(0)}^i) \times\right]\tilde{C}_b^i \delta\omega_{ib}^b \quad (70)$$

The differential equation of right error can be uniformly expressed as in (71).

$$\begin{bmatrix} \dot{\phi}_{ib}^i \\ J\dot{\rho}_{vib}^{iR} \\ J\dot{\rho}_{rib}^{iR} \end{bmatrix} = \begin{bmatrix} 0_{3\times 3} & 0_{3\times 3} & 0_{3\times 3} \\ (\tilde{\gamma}_{ib}^i \times) & 0_{3\times 3} & 0_{3\times 3} \\ 0_{3\times 3} & I_3 & 0_{3\times 3} \end{bmatrix} \begin{bmatrix} \phi_{ib}^i \\ J\rho_{vib}^{iR} \\ J\rho_{rib}^{iR} \end{bmatrix}$$
$$+ \begin{bmatrix} -\tilde{C}_b^i \delta\omega_{ib}^b \\ -\delta\gamma_{ib}^i - (\tilde{v}_{ib}^i \times)\tilde{C}_b^i \delta\omega_{ib}^b - \tilde{C}_b^i \delta f_{ib}^b \\ -\left[(\tilde{r}_{ib}^i - r_{ib(0)}^i) \times\right]\tilde{C}_b^i \delta\omega_{ib}^b \end{bmatrix} \quad (71)$$

According to Corollary 1 and (57), the differential equation of left error can be derived as in (72).

$$\frac{d}{dt}\eta_{ib}^{bL} = \eta_{ib}^{bL} W_1 - \tilde{W}_1 \eta_{ib}^{bL} - \left((\tilde{\chi}_{ib}^i)^{-1} \delta W_2 \tilde{\chi}_{ib}^i\right)\eta_{ib}^{bL}$$
$$= \eta_{ib}^{bL} \begin{bmatrix} (\omega_{ib}^b \times) & f_{ib}^b & 0_{3\times 1} \\ 0_{1\times 3} & 0 & 1 \\ 0_{1\times 3} & 0 & 0 \end{bmatrix} - \begin{bmatrix} (\tilde{\omega}_{ib}^b \times) & \tilde{f}_{ib}^b & 0_{3\times 1} \\ 0_{1\times 3} & 0 & 1 \\ 0_{1\times 3} & 0 & 0 \end{bmatrix}\eta_{ib}^{bL} \quad (72)$$
$$- \tilde{\chi}_{ib}^{i\;-1} \begin{bmatrix} 0_{3\times 3} & \delta\gamma_{ib}^i & 0_{3\times 1} \\ 0_{1\times 3} & 0 & 0 \\ 0_{1\times 3} & 0 & 0 \end{bmatrix} \tilde{\chi}_{ib}^i \eta_{ib}^{bL}$$

where

$$\eta_{ib}^{bL} = \begin{bmatrix} \exp(-\phi_{ib}^b \times) & J\rho_{vib}^{bL} & J\rho_{rib}^{bL} \\ 0_{1\times 3} & 1 & 0 \\ 0_{1\times 3} & 0 & 1 \end{bmatrix}, \tilde{\chi}_{ib}^i = \begin{bmatrix} \tilde{C}_b^i & \tilde{v}_{ib}^i & \tilde{r}_{ib}^i - r_{ib(0)}^i \\ 0_{1\times 3} & 1 & 0 \\ 0_{1\times 3} & 0 & 1 \end{bmatrix}$$
$$(73)$$

Rewrite (72) and perform a first-order approximation yields:

$$J\dot{\rho}_{vib}^{iL} \approx -\delta f_{ib}^b + (\tilde{f}_{ib}^b \times)\phi_{ib}^b - (\tilde{\omega}_{ib}^b \times)J\rho_{vib}^{iL} + \tilde{C}_i^b \delta\gamma_{ib}^i$$
$$J\dot{\rho}_{rib}^{bL} = J\rho_{vib}^{bL} - (\tilde{\omega}_{ib}^b \times)J\rho_{rib}^{bL} \quad (74)$$

The differential equation of left error can be uniformly expressed as in (75).

$$\begin{bmatrix} \dot{\phi}_{ib}^b \\ J\dot{\rho}_{vib}^{iL} \\ J\dot{\rho}_{rib}^{bL} \end{bmatrix} = \begin{bmatrix} -(\tilde{\omega}_{ib}^b \times) & 0_{3\times 3} & 0_{3\times 3} \\ (\tilde{f}_{ib}^b \times) & -(\tilde{\omega}_{ib}^b \times) & 0_{3\times 3} \\ 0_{3\times 3} & I_3 & -(\tilde{\omega}_{ib}^b \times) \end{bmatrix} \begin{bmatrix} \phi_{ib}^b \\ J\rho_{vib}^{iL} \\ J\rho_{rib}^{bL} \end{bmatrix}$$
$$+ \begin{bmatrix} \delta\omega_{ib}^b \\ -\delta f_{ib}^b + \tilde{C}_i^b \delta\gamma_{ib}^i \\ 0_{3\times 1} \end{bmatrix} \quad (75)$$

### 2) e-frame navigation error state and its derivative

The earth frame is taken as the projection frame and the reference frame for position, while the inertial frame is used as the reference frame for velocity. The attitude and position are still described with respect to the earth frame. The right error and left error of *e*-frame is given in (76).

$$\xi_{ieb}^{eR} = \begin{bmatrix} \phi_{eb}^e \\ J\rho_{vib}^{eR} \\ J\rho_{reb}^{eR} \end{bmatrix} \approx \begin{bmatrix} \phi_{eb}^e \\ -\delta v_{ib}^e + \left[(\tilde{v}_{ib}^e - (\omega_{ie}^e \times)r_{eb(0)}^e) \times\right]\phi_{eb}^e \\ -\delta r_{eb}^e + \left[(\tilde{r}_{eb}^e - r_{eb(0)}^e) \times\right]\phi_{eb}^e \end{bmatrix} \quad (76)$$

$$\xi_{ieb}^{bL} = \begin{bmatrix} \phi_{eb}^b \\ J\rho_{vib}^{bL} \\ J\rho_{reb}^{bL} \end{bmatrix} \approx \begin{bmatrix} \phi_{eb}^b \\ -\delta v_{ib}^b \\ -\delta r_{eb}^b \end{bmatrix}$$

The earth frame attitude error differential equation is given in (77).

$$\dot{\phi}_{eb}^e \approx -(\omega_{ie}^e \times)\phi_{eb}^e - \tilde{C}_b^e \delta\omega_{ib}^b \quad (77)$$

Similarly, the body frame counterpart is given in (78).

$$\dot{\phi}_{eb}^b \approx -(\tilde{\omega}_{ib}^b \times)\phi_{eb}^b + \delta\omega_{ib}^b \quad (78)$$

According to Corollary 1 and (55), the differential equation of right error can be derived as in (79).

$$\frac{d}{dt}\eta_{ieb}^{eR} = W_2 \eta_{ieb}^{eR} - \eta_{ieb}^{eR} \tilde{W}_2 - \eta_{ieb}^{eR}\left(\tilde{\chi}_{ieb}^e \delta W_1 (\tilde{\chi}_{ieb}^e)^{-1}\right)$$
$$= \begin{bmatrix} -(\omega_{ie}^e \times) & \gamma_{ib}^e - (\omega_{ie}^e \times)^2 r_{eb(0)}^e & 0_{3\times 1} \\ 0_{1\times 3} & 0 & -1 \\ 0_{1\times 3} & 0 & 0 \end{bmatrix}\eta_{ieb}^{eR}$$
$$- \eta_{ieb}^{eR}\begin{bmatrix} -(\omega_{ie}^e \times) & \tilde{\gamma}_{ib}^e - (\omega_{ie}^e \times)^2 r_{eb(0)}^e & -(\omega_{ie}^e \times)r_{eb(0)}^e \\ 0_{1\times 3} & 0 & -1 \\ 0_{1\times 3} & 0 & 0 \end{bmatrix} \quad (79)$$
$$- \eta_{ieb}^{eR}\tilde{\chi}_{ieb}^e \cdot \begin{bmatrix} (\delta\omega_{ib}^b \times) & \delta f_{ib}^b & 0_{3\times 1} \\ 0_{1\times 3} & 0 & 0 \\ 0_{1\times 3} & 0 & 0 \end{bmatrix} \tilde{\chi}_{ieb}^{e\;-1}$$

where

$$\boldsymbol{\eta}_{ieb}^{eR} = \begin{bmatrix} \exp(\boldsymbol{\phi}_{eb}^{e} \times) & \boldsymbol{J\rho}_{vib}^{eR} & \boldsymbol{J\rho}_{reb}^{eR} \\ \mathbf{0}_{1\times 3} & 1 & 0 \\ \mathbf{0}_{1\times 3} & 0 & 1 \end{bmatrix}$$

$$\tilde{\boldsymbol{\chi}}_{ieb}^{e} = \begin{bmatrix} \tilde{\boldsymbol{C}}_{b}^{e} & \tilde{\boldsymbol{v}}_{ib}^{e} - (\boldsymbol{\omega}_{ie}^{e} \times)\boldsymbol{r}_{eb(0)}^{e} & \tilde{\boldsymbol{r}}_{eb}^{e} - \boldsymbol{r}_{eb(0)}^{e} \\ \mathbf{0}_{1\times 3} & 1 & 0 \\ \mathbf{0}_{1\times 3} & 0 & 1 \end{bmatrix} \quad (80)$$

Rewrite (79) to a vector form and perform a first-order approximation we obtain (81).

$$\boldsymbol{J\dot{\rho}}_{vib}^{eR} = \left[\tilde{\boldsymbol{\gamma}}_{ib}^{e} - (\boldsymbol{\omega}_{ie}^{e} \times)^{2} \boldsymbol{r}_{eb(0)}^{e}\right] \times \boldsymbol{\phi}_{eb}^{e} - (\boldsymbol{\omega}_{ie}^{e} \times)\boldsymbol{J\rho}_{vib}^{eR}$$
$$- \left[(\tilde{\boldsymbol{v}}_{ib}^{e} - (\boldsymbol{\omega}_{ie}^{e} \times)\boldsymbol{r}_{eb(0)}^{e}) \times\right]\tilde{\boldsymbol{C}}_{b}^{e}\delta\boldsymbol{\omega}_{ib}^{b} - \tilde{\boldsymbol{C}}_{b}^{e}\delta\boldsymbol{f}_{ib}^{b} - \delta\boldsymbol{\gamma}_{ib}^{e} \quad (81)$$

$$\boldsymbol{J\dot{\rho}}_{reb}^{eR} = \boldsymbol{J\rho}_{vib}^{eR} - (\boldsymbol{\omega}_{ie}^{e} \times)\boldsymbol{J\rho}_{reb}^{eR} - \left[(\tilde{\boldsymbol{r}}_{eb}^{e} - \boldsymbol{r}_{eb(0)}^{e}) \times\right]\tilde{\boldsymbol{C}}_{b}^{e}\delta\boldsymbol{\omega}_{ib}^{b}$$

The differential equation of right error can be uniformly expressed as in (71).

$$\begin{bmatrix} \dot{\boldsymbol{\phi}}_{eb}^{e} \\ \boldsymbol{J\dot{\rho}}_{vib}^{eR} \\ \boldsymbol{J\dot{\rho}}_{reb}^{eR} \end{bmatrix} = \begin{bmatrix} -(\boldsymbol{\omega}_{ie}^{e}\times) & \mathbf{0}_{3\times 3} & \mathbf{0}_{3\times 3} \\ \left[(\tilde{\boldsymbol{\gamma}}_{ib}^{e} - (\boldsymbol{\omega}_{ie}^{e} \times)^{2}\boldsymbol{r}_{eb(0)}^{e})\times\right] & -(\boldsymbol{\omega}_{ie}^{e}\times) & \mathbf{0}_{3\times 3} \\ \mathbf{0}_{3\times 3} & \boldsymbol{I}_{3} & -(\boldsymbol{\omega}_{ie}^{e}\times) \end{bmatrix} \begin{bmatrix} \boldsymbol{\phi}_{eb}^{e} \\ \boldsymbol{J\rho}_{vib}^{eR} \\ \boldsymbol{J\rho}_{reb}^{eR} \end{bmatrix}$$
$$+ \begin{bmatrix} -\tilde{\boldsymbol{C}}_{b}^{e}\delta\boldsymbol{\omega}_{ib}^{b} \\ -\delta\boldsymbol{\gamma}_{ib}^{e} - \left[(\tilde{\boldsymbol{v}}_{ib}^{e} - (\boldsymbol{\omega}_{ie}^{e} \times)\boldsymbol{r}_{eb(0)}^{e})\times\right]\tilde{\boldsymbol{C}}_{b}^{e}\delta\boldsymbol{\omega}_{ib}^{b} - \tilde{\boldsymbol{C}}_{b}^{e}\delta\boldsymbol{f}_{ib}^{b} \\ -\left[(\tilde{\boldsymbol{r}}_{eb}^{e} - \boldsymbol{r}_{eb(0)}^{e})\times\right]\tilde{\boldsymbol{C}}_{b}^{e}\delta\boldsymbol{\omega}_{ib}^{b} \end{bmatrix} \quad (82)$$

According to Corollary 1 and (57), the differential equation of left error can be derived as in (83).

$$\frac{d}{dt}\boldsymbol{\eta}_{ieb}^{bL} = \boldsymbol{\eta}_{ieb}^{bL}\boldsymbol{W}_{1} - \tilde{\boldsymbol{W}}_{1}\boldsymbol{\eta}_{ieb}^{bL} - \left((\tilde{\boldsymbol{\chi}}_{ieb}^{e})^{-1}\delta\boldsymbol{W}_{2}\tilde{\boldsymbol{\chi}}_{ieb}^{e}\right)\boldsymbol{\eta}_{ieb}^{bL}$$

$$= \boldsymbol{\eta}_{ieb}^{bL}\begin{bmatrix} (\boldsymbol{\omega}_{ib}^{b}\times) & \boldsymbol{f}_{ib}^{b} & \mathbf{0}_{3\times 1} \\ \mathbf{0}_{1\times 3} & 0 & 1 \\ \mathbf{0}_{1\times 3} & 0 & 0 \end{bmatrix} - \begin{bmatrix} (\tilde{\boldsymbol{\omega}}_{ib}^{b}\times) & \tilde{\boldsymbol{f}}_{ib}^{b} & \mathbf{0}_{3\times 1} \\ \mathbf{0}_{1\times 3} & 0 & 1 \\ \mathbf{0}_{1\times 3} & 0 & 0 \end{bmatrix}\boldsymbol{\eta}_{ieb}^{bL} \quad (83)$$
$$- \tilde{\boldsymbol{\chi}}_{ieb}^{e\,-1}\begin{bmatrix} \mathbf{0}_{3\times 3} & \delta\boldsymbol{\gamma}_{ib}^{e} & \mathbf{0}_{3\times 1} \\ \mathbf{0}_{1\times 3} & 0 & 0 \\ \mathbf{0}_{1\times 3} & 0 & 0 \end{bmatrix}\tilde{\boldsymbol{\chi}}_{ieb}^{e}\boldsymbol{\eta}_{ieb}^{bL}$$

where

$$\boldsymbol{\eta}_{ib}^{bL} = \begin{bmatrix} \exp(-\boldsymbol{\phi}_{eb}^{b}\times) & \boldsymbol{J\rho}_{vieb}^{bL} & \boldsymbol{J\rho}_{reb}^{bL} \\ \mathbf{0}_{1\times 3} & 1 & 0 \\ \mathbf{0}_{1\times 3} & 0 & 1 \end{bmatrix}$$

$$\tilde{\boldsymbol{\chi}}_{ieb}^{i} = \begin{bmatrix} \tilde{\boldsymbol{C}}_{b}^{e} & \tilde{\boldsymbol{v}}_{ib}^{e} - (\boldsymbol{\omega}_{ie}^{e}\times)\boldsymbol{r}_{eb(0)}^{e} & \tilde{\boldsymbol{r}}_{eb}^{e} - \boldsymbol{r}_{eb(0)}^{e} \\ \mathbf{0}_{1\times 3} & 1 & 0 \\ \mathbf{0}_{1\times 3} & 0 & 1 \end{bmatrix} \quad (84)$$

Rewrite (83) to a vector form and perform a first-order approximation we obtain (85).

$$\boldsymbol{J\dot{\rho}}_{vieb}^{bL} = (\tilde{\boldsymbol{f}}_{ib}^{b}\times)\boldsymbol{\phi}_{eb}^{b} - (\tilde{\boldsymbol{\omega}}_{ib}^{b}\times)\boldsymbol{J\rho}_{vieb}^{bL} - \delta\boldsymbol{f}_{ib}^{b} + \tilde{\boldsymbol{C}}_{e}^{b}\delta\boldsymbol{\gamma}_{ib}^{e}$$
$$\boldsymbol{J\dot{\rho}}_{reb}^{bL} = \boldsymbol{J\rho}_{vib}^{bL} - (\tilde{\boldsymbol{\omega}}_{ib}^{b}\times)\boldsymbol{J\rho}_{reb}^{bL} \quad (85)$$

The differential equation of left error can be uniformly expressed as in (86).

$$\begin{bmatrix} \dot{\boldsymbol{\phi}}_{eb}^{b} \\ \boldsymbol{J\dot{\rho}}_{vieb}^{bL} \\ \boldsymbol{J\dot{\rho}}_{reb}^{bL} \end{bmatrix} = \begin{bmatrix} -(\tilde{\boldsymbol{\omega}}_{ib}^{b}\times) & \mathbf{0}_{3\times 3} & \mathbf{0}_{3\times 3} \\ (\tilde{\boldsymbol{f}}_{ib}^{b}\times) & -(\tilde{\boldsymbol{\omega}}_{ib}^{b}\times) & \mathbf{0}_{3\times 3} \\ \mathbf{0}_{3\times 3} & \boldsymbol{I}_{3} & -(\tilde{\boldsymbol{\omega}}_{ib}^{b}\times) \end{bmatrix}\begin{bmatrix} \boldsymbol{\phi}_{eb}^{b} \\ \boldsymbol{J\rho}_{vieb}^{bL} \\ \boldsymbol{J\rho}_{reb}^{bL} \end{bmatrix}$$
$$+ \begin{bmatrix} \delta\boldsymbol{\omega}_{ib}^{b} \\ -\delta\boldsymbol{f}_{ib}^{b} + \tilde{\boldsymbol{C}}_{e}^{b}\delta\boldsymbol{\gamma}_{ib}^{e} \\ \mathbf{0}_{3\times 1} \end{bmatrix} \quad (86)$$

### 3) w-frame navigation error state and its derivative

The $w$-frame is taken as the projection frame and the reference frame for position, while the Earth-centered inertial frame i is used as the reference frame for velocity. The attitude and position are still described relative to the world frame. The attitude and position are still described with respect to the world frame. The right error and left error of $w$-frame is given in (87).

$$\boldsymbol{\xi}_{iwb}^{wR} = \begin{bmatrix} \boldsymbol{\phi}_{wb}^{w} \\ \boldsymbol{J\rho}_{vib}^{wR} \\ \boldsymbol{J\rho}_{rwb}^{wR} \end{bmatrix} \approx \begin{bmatrix} \boldsymbol{\phi}_{wb}^{w} \\ -\delta\boldsymbol{v}_{ib}^{w} + \left[(\tilde{\boldsymbol{v}}_{ib}^{w} - (\boldsymbol{\omega}_{ie}^{w}\times)(\boldsymbol{r}_{ew}^{w} + \boldsymbol{r}_{wb(0)}^{w}))\times\right]\boldsymbol{\phi}_{wb}^{w} \\ -\delta\boldsymbol{r}_{wb}^{w} + \left[(\tilde{\boldsymbol{r}}_{wb}^{w} - \boldsymbol{r}_{wb(0)}^{w})\times\right]\boldsymbol{\phi}_{wb}^{w} \end{bmatrix}$$

$$\boldsymbol{\xi}_{iwb}^{bL} = \begin{bmatrix} \boldsymbol{\phi}_{wb}^{b} \\ \boldsymbol{J\rho}_{vib}^{bL} \\ \boldsymbol{J\rho}_{rwb}^{bL} \end{bmatrix} \approx \begin{bmatrix} \boldsymbol{\phi}_{wb}^{b} \\ -\delta\boldsymbol{v}_{ib}^{b} \\ -\delta\boldsymbol{r}_{wb}^{b} \end{bmatrix} \quad (87)$$

The attitude error differential equation is given.

$$\dot{\boldsymbol{\phi}}_{wb}^{w} \approx -(\boldsymbol{\omega}_{ie}^{w}\times)\boldsymbol{\phi}_{wb}^{w} - \tilde{\boldsymbol{C}}_{b}^{w}\delta\boldsymbol{\omega}_{ib}^{b} \quad (88)$$

$$\dot{\boldsymbol{\phi}}_{wb}^{b} \approx -(\tilde{\boldsymbol{\omega}}_{ib}^{b}\times)\boldsymbol{\phi}_{wb}^{b} + \delta\boldsymbol{\omega}_{ib}^{b} \quad (89)$$

According to Corollary 1 and (55), the differential equation of right error can be derived as in (90).

$$\frac{d}{dt}\boldsymbol{\eta}_{iwb}^{wR} = \boldsymbol{W}_{2}\boldsymbol{\eta}_{iwb}^{wR} - \boldsymbol{\eta}_{iwb}^{wR}\tilde{\boldsymbol{W}}_{2} - \boldsymbol{\eta}_{iwb}^{wR}\left(\tilde{\boldsymbol{\chi}}_{iwb}^{w}\delta\boldsymbol{W}_{1}(\tilde{\boldsymbol{\chi}}_{iwb}^{w})^{-1}\right)$$

$$= \boldsymbol{\eta}_{iwb}^{wR}\begin{bmatrix} \exp(\boldsymbol{\phi}_{wb}^{w}\times) & \boldsymbol{J\rho}_{vib}^{wR} & \boldsymbol{J\rho}_{rwb}^{wR} \\ \mathbf{0}_{1\times 3} & 1 & 0 \\ \mathbf{0}_{1\times 3} & 0 & 1 \end{bmatrix}$$
$$- \begin{bmatrix} \exp(\boldsymbol{\phi}_{wb}^{w}\times) & \boldsymbol{J\rho}_{vib}^{wR} & \boldsymbol{J\rho}_{rwb}^{wR} \\ \mathbf{0}_{1\times 3} & 1 & 0 \\ \mathbf{0}_{1\times 3} & 0 & 1 \end{bmatrix}\boldsymbol{\eta}_{iwb}^{wR} \quad (90)$$
$$- \boldsymbol{\eta}_{iwb}^{wR}\tilde{\boldsymbol{\chi}}_{iwb}^{w}\cdot\begin{bmatrix} (\delta\boldsymbol{\omega}_{ib}^{b}\times) & \delta\boldsymbol{f}_{ib}^{b} & \mathbf{0}_{3\times 1} \\ \mathbf{0}_{1\times 3} & 0 & 0 \\ \mathbf{0}_{1\times 3} & 0 & 0 \end{bmatrix}\tilde{\boldsymbol{\chi}}_{iwb}^{w\,-1}$$

where

$$\boldsymbol{\eta}_{iwb}^{wR} = \begin{bmatrix} \exp(\boldsymbol{\phi}_{wb}^{w} \times) & \boldsymbol{J}\boldsymbol{\rho}_{vib}^{wR} & \boldsymbol{J}\boldsymbol{\rho}_{rwb}^{wR} \\ \boldsymbol{0}_{1\times 3} & 1 & 0 \\ \boldsymbol{0}_{1\times 3} & 0 & 1 \end{bmatrix}$$

$$\tilde{\boldsymbol{\chi}}_{iwb}^{e} = \begin{bmatrix} \tilde{\boldsymbol{C}}_{b}^{w} & \tilde{\boldsymbol{v}}_{ib}^{w} - (\boldsymbol{\omega}_{ie}^{w} \times)(\boldsymbol{r}_{ew}^{w} + \boldsymbol{r}_{wb(0)}^{w}) & \tilde{\boldsymbol{r}}_{wb}^{w} - \boldsymbol{r}_{wb(0)}^{w} \\ \boldsymbol{0}_{1\times 3} & 1 & 0 \\ \boldsymbol{0}_{1\times 3} & 0 & 1 \end{bmatrix} \quad (91)$$

Rewrite (90) to a vector form and perform a first-order approximation we obtain (92).

$$\boldsymbol{J}\dot{\boldsymbol{\rho}}_{vib}^{wR} = \left[ \tilde{\boldsymbol{\gamma}}_{ib}^{w} - (\boldsymbol{\omega}_{ie}^{w} \times)^{2} (\boldsymbol{r}_{ew}^{w} + \boldsymbol{r}_{wb(0)}^{w}) \right] \boldsymbol{\phi}_{wb}^{w} - (\boldsymbol{\omega}_{ie}^{w} \times) \boldsymbol{J}\boldsymbol{\rho}_{vib}^{wR} - \delta \boldsymbol{\gamma}_{ib}^{w}$$
$$- \left[ (\tilde{\boldsymbol{v}}_{ib}^{w} - (\boldsymbol{\omega}_{ie}^{w} \times)(\boldsymbol{r}_{ew}^{w} + \boldsymbol{r}_{wb(0)}^{w})) \times \right] \tilde{\boldsymbol{C}}_{b}^{w} \delta \boldsymbol{\omega}_{ib}^{b} - \tilde{\boldsymbol{C}}_{b}^{w} \delta \boldsymbol{f}_{ib}^{b} \quad (92)$$
$$\boldsymbol{J}\dot{\boldsymbol{\rho}}_{rwb}^{wR} = \boldsymbol{J}\boldsymbol{\rho}_{vib}^{wR} - (\boldsymbol{\omega}_{ie}^{w} \times) \boldsymbol{J}\boldsymbol{\rho}_{rwb}^{wR} - \left[ (\tilde{\boldsymbol{r}}_{wb}^{w} - \boldsymbol{r}_{wb(0)}^{w}) \times \right] \tilde{\boldsymbol{C}}_{b}^{w} \delta \boldsymbol{\omega}_{ib}^{b}$$

The differential equation of right error can be uniformly expressed as in (93):

$$\begin{bmatrix} \dot{\boldsymbol{\phi}}_{wb}^{w} \\ \boldsymbol{J}\dot{\boldsymbol{\rho}}_{vib}^{wR} \\ \boldsymbol{J}\dot{\boldsymbol{\rho}}_{rwb}^{wR} \end{bmatrix} = \begin{bmatrix} -(\boldsymbol{\omega}_{ie}^{w} \times) & \boldsymbol{0}_{3\times 3} & \boldsymbol{0}_{3\times 3} \\ (\boldsymbol{\gamma}_{ib}^{w} - (\boldsymbol{\omega}_{ie}^{w} \times)^{2}(\boldsymbol{r}_{ew}^{w} + \boldsymbol{r}_{wb(0)}^{w})) \times & -(\boldsymbol{\omega}_{ie}^{w} \times) & \boldsymbol{0}_{3\times 3} \\ \boldsymbol{0}_{3\times 3} & \boldsymbol{I}_{3} & -(\boldsymbol{\omega}_{ie}^{w} \times) \end{bmatrix} \begin{bmatrix} \boldsymbol{\phi}_{wb}^{w} \\ \boldsymbol{J}\boldsymbol{\rho}_{vib}^{wR} \\ \boldsymbol{J}\boldsymbol{\rho}_{rwb}^{wR} \end{bmatrix}$$
$$+ \begin{bmatrix} -\tilde{\boldsymbol{C}}_{b}^{w} \delta \boldsymbol{\omega}_{ib}^{b} \\ -\delta \boldsymbol{\gamma}_{ib}^{w} - \left[ (\tilde{\boldsymbol{v}}_{ib}^{w} - (\boldsymbol{\omega}_{ie}^{w} \times)(\boldsymbol{r}_{ew}^{w} + \boldsymbol{r}_{wb(0)}^{w})) \times \right] \tilde{\boldsymbol{C}}_{b}^{w} \delta \boldsymbol{\omega}_{ib}^{b} - \tilde{\boldsymbol{C}}_{b}^{w} \delta \boldsymbol{f}_{ib}^{b} \\ - \left[ (\tilde{\boldsymbol{r}}_{wb}^{w} - \boldsymbol{r}_{wb(0)}^{w}) \times \right] \tilde{\boldsymbol{C}}_{b}^{w} \delta \boldsymbol{\omega}_{ib}^{b} \end{bmatrix}$$
(93)

According to Corollary 1 and (57), the differential equation of left error can be derived as in (94).

$$\frac{d}{dt} \boldsymbol{\eta}_{iwb}^{bL} = \boldsymbol{\eta}_{iwb}^{bL} \boldsymbol{W}_{1} - \tilde{\boldsymbol{W}}_{1} \boldsymbol{\eta}_{iwb}^{bL} - \left( (\tilde{\boldsymbol{\chi}}_{iwb}^{w})^{-1} \delta \boldsymbol{W}_{2} \tilde{\boldsymbol{\chi}}_{iwb}^{w} \right) \boldsymbol{\eta}_{iwb}^{bL}$$
$$= \boldsymbol{\eta}_{iwb}^{bL} \begin{bmatrix} (\boldsymbol{\omega}_{ib}^{b} \times) & \boldsymbol{f}_{ib}^{b} & \boldsymbol{0}_{3\times 1} \\ \boldsymbol{0}_{1\times 3} & 0 & 1 \\ \boldsymbol{0}_{1\times 3} & 0 & 0 \end{bmatrix} - \begin{bmatrix} (\tilde{\boldsymbol{\omega}}_{ib}^{b} \times) & \tilde{\boldsymbol{f}}_{ib}^{b} & \boldsymbol{0}_{3\times 1} \\ \boldsymbol{0}_{1\times 3} & 0 & 1 \\ \boldsymbol{0}_{1\times 3} & 0 & 0 \end{bmatrix} \boldsymbol{\eta}_{iwb}^{bL} \quad (94)$$
$$- \tilde{\boldsymbol{\chi}}_{iwb}^{w \; -1} \begin{bmatrix} \boldsymbol{0}_{3\times 3} & \delta \boldsymbol{\gamma}_{ib}^{w} & \boldsymbol{0}_{3\times 1} \\ \boldsymbol{0}_{1\times 3} & 0 & 0 \\ \boldsymbol{0}_{1\times 3} & 0 & 0 \end{bmatrix} \cdot \tilde{\boldsymbol{\chi}}_{iwb}^{w} \boldsymbol{\eta}_{iwb}^{bL}$$

where

$$\boldsymbol{\eta}_{iwb}^{bL} = \begin{bmatrix} \exp(-\boldsymbol{\phi}_{wb}^{b} \times) & \boldsymbol{J}\boldsymbol{\rho}_{vib}^{bL} & \boldsymbol{J}\boldsymbol{\rho}_{rwb}^{bL} \\ \boldsymbol{0}_{1\times 3} & 1 & 0 \\ \boldsymbol{0}_{1\times 3} & 0 & 1 \end{bmatrix}$$

$$\tilde{\boldsymbol{\chi}}_{iwb}^{w} = \begin{bmatrix} \tilde{\boldsymbol{C}}_{b}^{w} & \tilde{\boldsymbol{v}}_{ib}^{w} - (\boldsymbol{\omega}_{ie}^{w} \times) \boldsymbol{r}_{wb(0)}^{w} & \tilde{\boldsymbol{r}}_{wb}^{w} - \boldsymbol{r}_{wb(0)}^{w} \\ \boldsymbol{0}_{1\times 3} & 1 & 0 \\ \boldsymbol{0}_{1\times 3} & 0 & 1 \end{bmatrix} \quad (95)$$

Rewrite (94) to a vector form and perform a first-order approximation we obtain (96).

$$\boldsymbol{J}\dot{\boldsymbol{\rho}}_{vib}^{bL} \approx \left( \tilde{\boldsymbol{f}}_{ib}^{b} \times \right) \boldsymbol{\phi}_{wb}^{b} - (\tilde{\boldsymbol{\omega}}_{ib}^{b} \times) \boldsymbol{J}\boldsymbol{\rho}_{vib}^{bL} - \delta \boldsymbol{f}_{ib}^{b} + \tilde{\boldsymbol{C}}_{w}^{b} \delta \boldsymbol{\gamma}_{ib}^{w}$$
$$\boldsymbol{J}\dot{\boldsymbol{\rho}}_{rwb}^{bL} = \boldsymbol{J}\boldsymbol{\rho}_{vib}^{bL} - (\tilde{\boldsymbol{\omega}}_{ib}^{b} \times) \boldsymbol{J}\boldsymbol{\rho}_{rwb}^{bL} \quad (96)$$

The differential equation of left error can be uniformly expressed as in (97).

$$\begin{bmatrix} \dot{\boldsymbol{\phi}}_{wb}^{b} \\ \boldsymbol{J}\dot{\boldsymbol{\rho}}_{vib}^{bL} \\ \boldsymbol{J}\dot{\boldsymbol{\rho}}_{rwb}^{bL} \end{bmatrix} = \begin{bmatrix} -(\tilde{\boldsymbol{\omega}}_{ib}^{b} \times) & \boldsymbol{0}_{3\times 3} & \boldsymbol{0}_{3\times 3} \\ (\tilde{\boldsymbol{f}}_{ib}^{b} \times) & -(\tilde{\boldsymbol{\omega}}_{ib}^{b} \times) & \boldsymbol{0}_{3\times 3} \\ \boldsymbol{0}_{3\times 3} & \boldsymbol{I}_{3} & -(\tilde{\boldsymbol{\omega}}_{ib}^{b} \times) \end{bmatrix} \begin{bmatrix} \boldsymbol{\phi}_{wb}^{b} \\ \boldsymbol{J}\boldsymbol{\rho}_{vib}^{bL} \\ \boldsymbol{J}\boldsymbol{\rho}_{rwb}^{bL} \end{bmatrix}$$
$$+ \begin{bmatrix} \delta \boldsymbol{\omega}_{ib}^{b} \\ -\delta \boldsymbol{f}_{ib}^{b} + \tilde{\boldsymbol{C}}_{w}^{b} \delta \boldsymbol{\gamma}_{ib}^{w} \\ \boldsymbol{0}_{3\times 1} \end{bmatrix} \quad (97)$$

It can be seen that the proposed navigation models under non-inertial frame both have approximate autonomous properties after velocity replacement.

**C. Measurement equation of the SINS/ODO integration**

This section derives the improved $SE_2(3)$ group based SINS/ODO integrated navigation measurement equations.

The observation is the velocity calculated by the SINS minus the front velocity measured by the odometer. It can be represented as

$$\delta z_v = \boldsymbol{v}_{INS}^{b} - \boldsymbol{v}_{ODO}^{b} \quad (98)$$

The measurement equation under different frames is given below.

**1) *i*-frame measurement equations**

The measurement equation of left error in *i*-frame is derived as (99).

$$\delta z_v = \boldsymbol{v}_{INS}^{b} - \boldsymbol{v}_{ODO}^{b} = \tilde{\boldsymbol{C}}_{i}^{b} \tilde{\boldsymbol{v}}_{eb}^{i} - \boldsymbol{v}_{ODO}^{b} = \boldsymbol{C}_{b}^{b'} \boldsymbol{C}_{i}^{b} \tilde{\boldsymbol{v}}_{eb}^{i} - \boldsymbol{v}_{ODO}^{b}$$
$$= (\boldsymbol{I} - \boldsymbol{\phi}_{ib}^{b} \times) \boldsymbol{C}_{i}^{b} (\boldsymbol{v}_{eb}^{i} + \delta \boldsymbol{v}_{ib}^{i} - \boldsymbol{\omega}_{ie}^{i} \times \delta \boldsymbol{r}_{ib}^{i}) - \boldsymbol{v}_{ODO}^{b}$$
$$= \boldsymbol{C}_{i}^{b} (\boldsymbol{v}_{eb}^{i} + \delta \boldsymbol{v}_{ib}^{i} - \boldsymbol{\omega}_{ie}^{i} \times \delta \boldsymbol{r}_{ib}^{i}) - (\boldsymbol{\phi}_{ib}^{b} \times) \boldsymbol{C}_{i}^{b} \boldsymbol{v}_{eb}^{i} - \boldsymbol{v}_{ODO}^{b} \quad (99)$$
$$= (\boldsymbol{v}_{eb}^{b} \times) \boldsymbol{\phi}_{ib}^{b} + \delta \boldsymbol{v}_{ib}^{b} - (\boldsymbol{\omega}_{ie}^{b} \times) \delta \boldsymbol{r}_{ib}^{b}$$

The measurement equation of right error in *i*-frame is derived as (100).

$$\delta z_v = \boldsymbol{v}_{INS}^{b} - \boldsymbol{v}_{ODO}^{b} = \tilde{\boldsymbol{C}}_{i}^{b} \tilde{\boldsymbol{v}}_{eb}^{i} - \boldsymbol{v}_{ODO}^{b} = \boldsymbol{C}_{i}^{b} \boldsymbol{C}_{i}^{i} \tilde{\boldsymbol{v}}_{eb}^{i} - \boldsymbol{v}_{ODO}^{b}$$
$$= \boldsymbol{C}_{i}^{b} (\boldsymbol{I} + \boldsymbol{\phi}_{ib}^{i} \times)(\boldsymbol{v}_{eb}^{i} + \delta \boldsymbol{v}_{ib}^{i} - \boldsymbol{\omega}_{ie}^{i} \times \delta \boldsymbol{r}_{ib}^{i}) - \boldsymbol{v}_{ODO}^{b}$$
$$= -\boldsymbol{C}_{i}^{b} (\boldsymbol{v}_{eb}^{i} \times) \boldsymbol{\phi}_{ib}^{i} + \boldsymbol{C}_{i}^{b} \delta \boldsymbol{v}_{ib}^{i} - \boldsymbol{C}_{i}^{b} (\boldsymbol{\omega}_{ie}^{i} \times) \delta \boldsymbol{r}_{ib}^{i}$$
$$= -\boldsymbol{C}_{i}^{b} ((\boldsymbol{v}_{ib}^{i} - (\boldsymbol{\omega}_{ie}^{i} \times \boldsymbol{r}_{ib}^{i})) \times) \boldsymbol{\phi}_{ib}^{i} + \boldsymbol{C}_{i}^{b} \delta \boldsymbol{v}_{ib}^{i} - \boldsymbol{C}_{i}^{b} (\boldsymbol{\omega}_{ie}^{i} \times) \delta \boldsymbol{r}_{ib}^{i} \quad (100)$$
$$= \boldsymbol{C}_{i}^{b} \begin{bmatrix} (\boldsymbol{\omega}_{ie}^{i} \times \boldsymbol{r}_{ib}^{i}) \times \\ -(\boldsymbol{\omega}_{ie}^{i} \times)((\boldsymbol{r}_{ib}^{i} - \boldsymbol{r}_{ib(0)}^{i}) \times) \end{bmatrix} \boldsymbol{\phi}_{ib}^{i} - \boldsymbol{C}_{i}^{b} \boldsymbol{J}\boldsymbol{\rho}_{vib}^{iR} + \boldsymbol{C}_{i}^{b} (\boldsymbol{\omega}_{ie}^{i} \times) \boldsymbol{J}\boldsymbol{\rho}_{rib}^{iR}$$

### 2) *e*-frame measurement equations

The measurement equation of left error in *e*-frame is derived as (101).

$$\delta z_v = \tilde{v}_{INS}^b - \tilde{v}_{ODO}^b = (I - \phi_{eb}^b \times) C_e^b (v_{eb}^e + \delta v_{eb}^e) - v_{ODO}^b \\ = v_{eb}^b \times \phi_{eb}^b + \delta v_{eb}^b \quad (101)$$

The measurement equation of right error in *e*-frame is derived as (102).

$$\delta z_v = \tilde{v}_{INS}^b - \tilde{v}_{ODO}^b = \tilde{C}_e^b (v_{eb}^e + \delta v_{eb}^e) - v_{ODO}^b \\ = C_e^b (I + \phi_{eb}^e \times)(v_{eb}^e + \delta v_{eb}^e) - v_{ODO}^b \\ = C_e^b (v_{eb}^e + \delta v_{eb}^e) - v_{ODO}^b + C_e^b (\phi_{eb}^e \times) v_{eb}^e \\ = -C_e^b J \rho_v^e \quad (102)$$

The measurement equation of proposed improved left error in *i*-frame is derived as (103).

$$\delta z_v = \tilde{v}_{INS}^b - \tilde{v}_{ODO}^b = \tilde{C}_{i_w}^b \tilde{v}_{eb}^{i_w} - v_{ODO}^b = C_b^{b'} C_{i_w}^b \tilde{v}_{eb}^{i_w} - v_{ODO}^b \\ = (I - \phi_{ib}^b \times) C_{i_w}^b (v_{eb}^{i_w} + \delta v_{eb}^{i_w} - \omega_{ie}^{i_w} \times \delta r_{eb}^{i_w}) - v_{ODO}^b \\ = (v_{eb}^b \times) \phi_{ib}^b + \delta v_{ib}^b - C_{i_w}^b (\omega_{ie}^{i_w} \times) C_b^{i_w} \delta r_{eb}^b \\ = (v_{eb}^b \times) \phi_{ib}^b + \delta v_{ib}^b - (\omega_{ie}^b \times) \delta r_{eb}^b \quad (103)$$

The measurement equation of proposed improved right error in *e*-frame is derived as (104).

$$\delta z_v = \tilde{v}_{INS}^b - \tilde{v}_{ODO}^b = -C_e^b J \rho_v^e \\ = -C_e^b J \rho_{vib}^{eR} + C_e^b (\omega_{ie}^e) \times J \rho_{rib}^{eR} \quad (104)$$

### 3) *w*-frame measurement equations

Similarly, the measurement equations are directly given below. It is a simple replacement of *e*-frame by *w*-frame.

$$\delta z_v = v_{wb}^b \times \phi_{wb}^b + \delta v_{wb}^b \quad (105)$$

$$\delta z_v = -C_w^b J \rho_v^w \quad (106)$$

$$\delta z_v = (v_{wb}^b \times) \phi_{wb}^b + \delta v_{ib}^b - (\omega_{ie}^b \times) \delta r_{wb}^b \quad (107)$$

$$\delta z_v = -C_w^b J \rho_{vib}^{wR} + C_w^b (\omega_{ie}^w) \times J \rho_{rib}^{wR} \quad (108)$$

## V. CONCLUSION

This paper discussed the autonomy property of state-of-art $SE_2(3)$ group based navigation model and pointed out that autonomy property is disturbed under non-inertial frame. This paper proposed a construction method to improve the autonomy property for non-inertial frame navigation model by replacing the non-inertial velocity with inertial velocity. The SINS/ODO integrated navigation measurement equations under different frames are derived. The Monte-Carlo and real-world experiment results are given in the counterpart of this paper.

## REFERENCES

[1] R. E. Kalman, "A New Approach to Linear Filtering and Prediction Problems." *Trans. ASME, J. Basic Eng.*, vol. Series D 82, pp. 35-45, Mar. 1960.

[2] R. E. Kalman, and R.S. Bucy, "New Results in Linear Filtering and Prediction Theory," *Trans. ASME, J. Basic Eng.*, vol. Series D 82, pp. 95-108, Mar. 1961.

[3] S. F. Schmidt, "The Kalman Filter: Its Recognition and Development for Aerospace Applications," *J. Guid. Control*, vol. 4, pp. 4-7, Jan.-Feb. 1981.

[4] E. J. Lefferts, F. L. Markley, and M. D. Shuster, "Kalman Filtering for Spacecraft Attitude Estimation," *J. Guid., Control, Dyn.*, vol. 5, no. 5, pp. 417–429, Sep. 1982.

[5] V. J. Aidala, "Kalman Filter Behavior in Bearings-Only Tracking Applications," *IEEE Trans. Aerosp. Electron. Syst.*, vol. AES-15, no. 1, pp. 29-39, Jan. 1979.

[6] Y. Huang, Y. Zhang, N. Li, Z. Wu and J. A. Chambers. "A novel robust student's t-based Kalman filter." *IEEE Trans. Aerosp. Electron. Syst.*, vol. 53, no. 3, pp. 1545-1554, June 2017.

[7] M. S. Grewal and A. P. Andrews, "Applications of Kalman Filtering in Aerospace 1960 to the Present," *IEEE Control Syst. Mag.*, vol. 30, no. 3, pp. 69-78, June 2010.

[8] F. L. Markley, "Attitude Error Representations for Kalman Filtering," *J. Guid., Control, Dyn.*, vol. 26, no. 2, pp. 311–317, Mar. 2003.

[9] H. A. Hashim, L. J. Brown and K. McIsaac, "Nonlinear Pose Filters on the Special Euclidean Group SE(3) With Guaranteed Transient and Steady-State Performance," *IEEE Trans. Syst., Man, Cybern.*, vol. 51, no. 5, pp. 2949-2962, May 2021.

[10] T. Qin, P. Li and S. Shen, "VINS-Mono: A Robust and Versatile Monocular Visual-Inertial State Estimator," *IEEE Trans. Robot.*, vol. 34, no. 4, pp. 1004-1020, Aug. 2018.

[11] M. Zhang, X. Zuo, Y. Chen, Y. Liu and M. Li, "Pose Estimation for Ground Robots: On Manifold Representation, Integration, Reparameterization, and Optimization," *IEEE Trans. Robot.*, vol. 37, no. 4, pp. 1081-1099, Aug. 2021.

[12] A. Manzanilla, S. Reyes, M. Garcia, D. Mercado and R. Lozano, "Autonomous Navigation for Unmanned Underwater Vehicles: Real-Time Experiments Using Computer Vision," *IEEE Robot. Autom. Lett.*, vol. 4, no. 2, pp. 1351-1356, Apr. 2019.

[13] J. L. Crassidis. and F. L. Markley, "Unscented Filtering for Spacecraft Attitude Estimation," *J. Guid., Control, Dyn.*, vol. 26, no. 4, pp. 536–542, May 2003.

[14] F. Gustafsson et al., "Particle filters for positioning, navigation, and tracking," *IEEE Trans. Signal Process.*, vol. 50, no. 2, pp. 425-437, Feb. 2002.

[15] A. Barrau. "Non-linear state error based extended Kalman filters with applications to navigation," Ph.D. dissertation, Mines Paris Tech, Paris, France, 2015.

[16] A. Barrau and S. Bonnabel, "The Invariant Extended Kalman Filter as a Stable Observer," *IEEE Trans. Autom. Control*, vol. 62, no. 4, pp. 1797-1812, Apr. 2017

[17] A. Barrau and S. Bonnabel, "The Geometry of Navigation Problems," *IEEE Trans. Autom. Control*, vol. 68, no. 2, pp. 689-704, Feb. 2023.

[18] M. Wang, J. Cui and W. Wu, "Left/Right Invariant Lie Group Error for SINS/GNSS Tightly Coupled Vehicular Integrated Navigation," *IEEE Trans. Veh. Technol.*, vol. 74, no. 6, pp. 8975-8988, June 2025.

[19] J. Cui, M. Wang, W. Wu and X. He. "Lie group based nonlinear state errors for MEMS-IMU/GNSS/magnetometer integrated navigation," *J. Navig.*, vol. 74, no. 4, pp. 887-900, Mar. 2021.

[20] J. Cui, M. Wang, W. Wu, R. Liu and W. Yang, "Enhanced LG-EKF Backtracking Framework for Body-Velocity-Aided Vehicular Integrated Navigation," *IEEE Trans. Veh. Technol.*, vol. 73, no. 10, pp. 14212-14223, Oct. 2024.

[21] L. Chang, H. Tang, G. Hu and J. Xu, "SINS/DVL Linear Initial Alignment Based on Lie Group SE3(3)," *IEEE Trans. Aerosp. Electron. Syst.*, vol. 59, no. 5, pp. 7203-7217, Oct. 2023.

[22] G. Hu, J. Geng, L. Chang, B. Gao and Y. Zhong, "Tightly Coupled SINS/DVL Based on Lie Group SE2(3) in Local-Level Frame," *IEEE Trans. Aerosp. Electron. Syst.*, [Early Access]. Available: https://ieeexplore.ieee.org/abstract/document/11098863, doi: 10.1109/TAES.2025.3593458.

[23] A. Fornasier, Y. Ng, R. Mahony and S. Weiss, "Equivariant Filter Design for Inertial Navigation Systems with Input Measurement Biases," 2022 *Proc. Int. Conf. Robot. Autom. (ICRA)*, Philadelphia, PA, USA, 2022, pp. 4333-4339.

[24] P. van Goor and R. Mahony, "EqVIO: An Equivariant Filter for Visual-Inertial Odometry," *IEEE Trans. Robot.*, vol. 39, no. 5, pp. 3567-3585, Oct. 2023.

[25] Y. Luo, C. Guo, & J. Liu, "Equivariant filtering framework for inertial-integrated navigation." *Satell. Navig.*, vol. 2, no. 30, pp. 1-17, Dec. 2021.

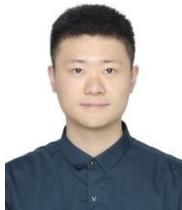

**Jiarui. Cui** received the B.S. degree from the Nanjing University of Aeronautics and Astronautics in 2018, the M.S. degree and Ph.D. degree from the National University of Defense Technology in 2020 and 2025. He is now an assistant researcher at Beijing Institute of Tracking and Telecommunications Technology. His research interests include inertial based multi-sensor integrated navigation and advanced state estimation algorithms.

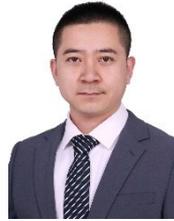

**Maosong Wang** received the B.S. degree from Harbin Engineering University in 2012, the M.S. degree from the National University of Defense Technology in 2014, and the Ph.D. degree from the National University of Defense Technology in 2018. From September 2016 to March 2018, he was a Visiting Student Researcher at the University of Calgary, Canada. Currently, he is an associated professor at the National University of Defense Technology. His research interests include inertial navigation algorithm, and multi-sensor integrated navigation theory and application.

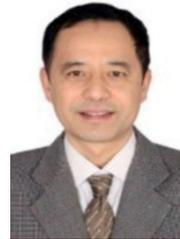

**Wenqi Wu** received the B.S. degree from Tianjin University in 1988, the M.S. degree from the National University of Defense Technology in 1991, and the Ph.D. degree from the National University of Defense Technology in 2002. Currently, he is a professor at the National University of Defense Technology. His research interests include GNSS, inertial, and multi-sensor integrated navigation theory and application.

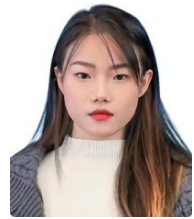

**Peiqi Li** received the B.S. degree from Harbin Engineering University in 2024.Currently,she is pursuing the degree of master in the National University of Defense Technology. Her research interests include inertial navigation algorithms, Lie group based Kalman filter theory and application.

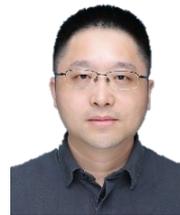

**Xianfei Pan** received the Ph.D. degree in control science and engineering from the National University of Defense Technology, Changsha, China, in 2008. Currently, he is a Professor of the College of Intelligence Science and Technology, National University of Defense Technology. His current research interests include multi-source integrated navigation, cooperative navigation and intelligent navigation.